\documentclass[twoside]{article}
\usepackage[accepted]{aistats2024}

\usepackage{amsmath,amsfonts,bm}

\def\eqref#1{equation~\ref{#1}}

\def\1{\bm{1}}

\def\vx{{\bm{x}}}

\def\mA{{\mathbf{A}}}

\def\mC{{\mathbf{C}}}

\def\mQ{{\mathbf{Q}}}

\def\mX{{\mathbf{X}}}

\DeclareMathAlphabet{\mathsfit}{\encodingdefault}{\sfdefault}{m}{sl}
\SetMathAlphabet{\mathsfit}{bold}{\encodingdefault}{\sfdefault}{bx}{n}

\def\targetgraph{\ensuremath{\mathcal{G}_T}}
\def\targetnode{\ensuremath{\mathcal{V}_T}}
\def\targetedge{\ensuremath{\mathcal{E}_T}}

\def\observedgraph{\ensuremath{\mathcal{G}_O}}
\def\observednode{\ensuremath{\mathcal{V}_O}}
\def\observededge{\ensuremath{\mathcal{E}_O}}
\def\missingedge{\ensuremath{\mathcal{E}_M}}
\def\additionaledge{\ensuremath{\mathcal{E}_A}}
\def\gengraph{\ensuremath{\mathcal{G}_G}}
\def\gennode{\ensuremath{\mathcal{V}_G}}
\def\genedge{\ensuremath{\mathcal{E}_G}}

\usepackage{hyperref}
\usepackage{url}
\usepackage{enumitem}
\usepackage{comment}
\usepackage{graphicx}
\usepackage{amssymb}
\usepackage[ruled]{algorithm2e}
\usepackage{caption}
\usepackage{subcaption}
\usepackage{dsfont}
\usepackage{xcolor,colortbl}
\usepackage{longtable}
\usepackage{wrapfig}
\usepackage{todonotes}
\usepackage{booktabs}
\usepackage{setspace}
\usepackage{multirow}
\usepackage{array}
\usepackage{cleveref}
\usepackage[utf8]{inputenc}

\SetCommentSty{mycommfont}
\SetKwComment{Comment}{$\triangleright$\ }{}

\newcommand\mypar[1]{\noindent \textbf{\textit{#1}}}
\newcommand\ours{\textbf{\texttt{SGDM}}\xspace}
\SetKwFor{While}{while}{}{end while}%
\usepackage{pifont}
\newcommand{\xmark}{\ding{55}}%
\hypersetup{
    colorlinks,
    linkcolor={red!50!black},
    citecolor={blue!50!black},
    urlcolor={blue!80!black}
}
\newcolumntype{H}{>{\setbox0=\hbox\bgroup}c<{\egroup}@{}}

\usepackage{color, colortbl}
\definecolor{Gray}{gray}{0.9}

\usepackage[round]{natbib}

\begin{document}
\pagenumbering{arabic}
\runningtitle{Leveraging Graph Diffusion Models for Network Refinement Tasks}

\runningauthor{Leveraging Graph Diffusion Models for Network Refinement Tasks}

\twocolumn[

\aistatstitle{Leveraging Graph Diffusion Models for Network Refinement Tasks}

\aistatsauthor{ Puja Trivedi \And Ryan Rossi \And David Arbour \And Tong Yu \And Franck Dernoncourt}
\aistatsaddress{ University of Michigan \And  Adobe Research \And Adobe Research\And Adobe Research\And Adobe Research} 
\aistatsauthor{Sungchul Kim \And Nedim Lipka \And Namyong Park \And Nesreen K. Ahmed \And Danai Koutra}
\aistatsaddress{Adobe\And  Adobe \And Meta \And  Intel Labs \And University of Michigan}]

\begin{abstract}
Most real-world networks are noisy and incomplete samples from an unknown target distribution. Refining them by correcting corruptions or inferring unobserved regions typically improves downstream performance. 
Inspired by the impressive generative capabilities that have been used to correct corruptions in \textit{images}, 
and the similarities between ``in-painting" and filling in missing nodes and edges conditioned on the observed graph, we propose a novel graph generative framework, \ours, which is based on subgraph diffusion. Our framework not only improves the scalability and fidelity of graph diffusion models,  
but also leverages the reverse process to perform novel, conditional generation tasks. In particular, through extensive empirical analysis and a set of novel metrics, we demonstrate that our proposed model effectively supports the following refinement tasks for partially observable networks: \texttt{(T1)}~denoising extraneous subgraphs, \texttt{(T2)}~expanding existing subgraphs and \texttt{(T3)}~performing ``style" transfer by regenerating a particular subgraph to match the characteristics of a different node or subgraph. 
\end{abstract}

\section{Introduction}
\begin{figure}[t]
    \centering
    \includegraphics[width=0.9\columnwidth]{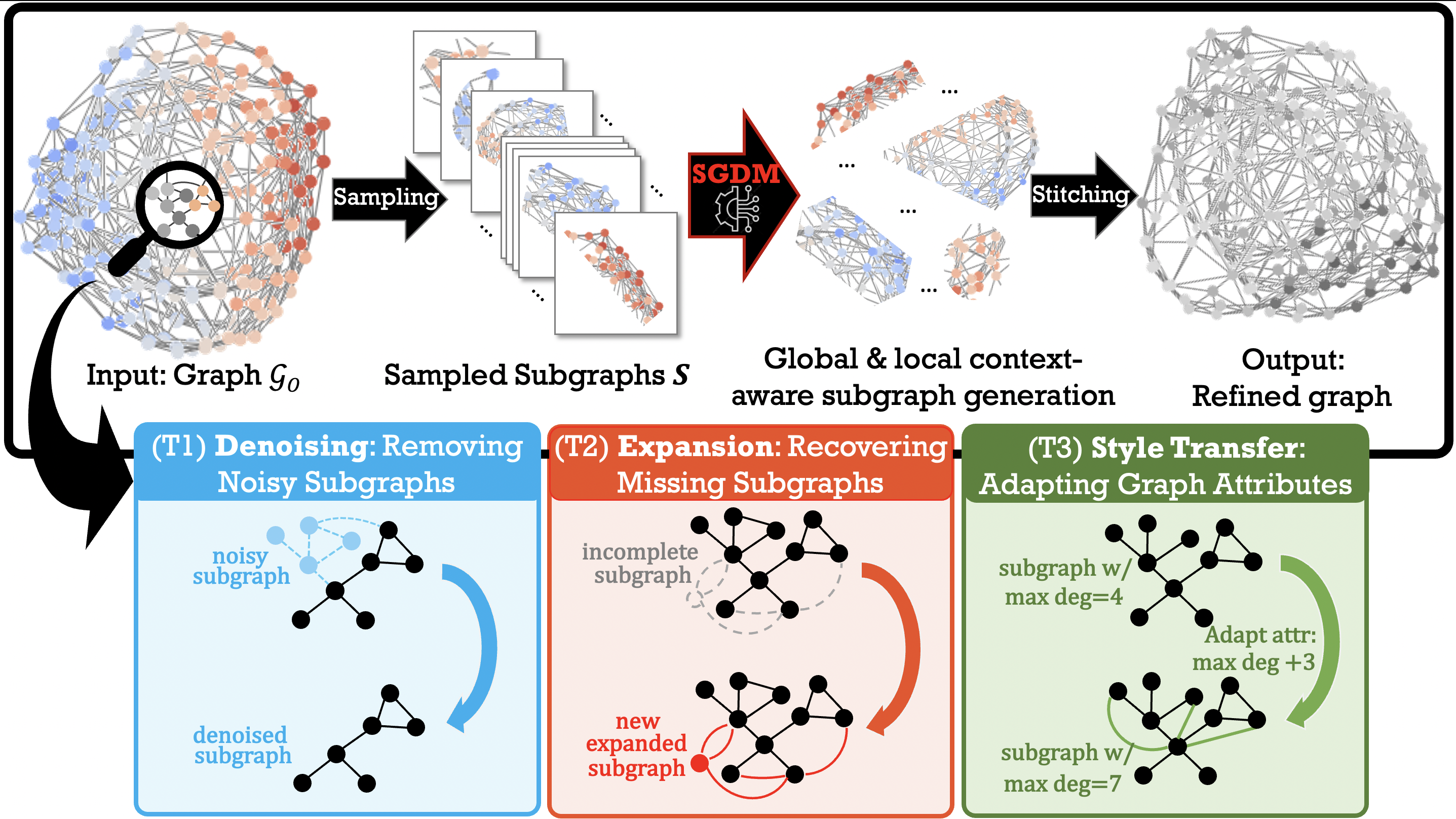}
    \caption{\textbf{Overview of Subgraph-Diffusion Models and Editing Tasks.} \ours uses subgraph sampling to flexibly support three useful editing tasks for refining a single, partially observable network.}
    \label{fig:overview}
\end{figure}
Real-world networks are typically partially observed and noisy~\citep{EliassiRad19_IncompletenessNetworks,Kim11_NetworkCompletion,Hanneke09_NetworkCompletionSurvey},  posing challenges in downstream tasks (e.g., node classification). 
Refining a corrupted graph can significantly improve downstream performance but includes a variety of editing tasks with different requirements such as: denoising (removing extraneous edges), expansion (recovering missing edges), and attribute editing (correcting erroneous or missing node attributes). 
Since the observed graph can be corrupted in different ways, it is important to jointly support all these editing tasks, but existing methods usually only support some of them. For instance, link prediction methods~\citep{Zhang18_LinkPred,Zhang20_RevistingLinkPred} are able to recover missing edges, but cannot remove noisy edges. Likewise, graph generative adversarial networks (GANs) \citep{Cao18_MolGAN,Liao19_GRAN,Wang18_GraphGAN,Martinkus22_SPECTRE} can be used to generate candidate subgraphs that maximize a given property by using RL~\citep{Cao18_MolGAN,Zhu22_GraphGenSurveyLog}, but it can be difficult to enforce that certain desired subgraphs are preserved in the generated graph (i.e., conditional generation).

Recently, in computer vision, large-scale diffusion models (DMs)~\citep{Ho20_DDPM,Hoogeboom23_simplediffusion,Dickstein15_Nonequilibrium,Ho22_CascadedDiffusion} have revolutionized image generation and correction of corrupted images due to (1)~their ability to generate extremely high-quality samples while capturing many modes of the underlying data distribution, (2)~their ability to easily support powerful conditional generation and editing tasks, such as in-painting, colorization, panorama generation and artifact removal~\citep{Lugmayr22_RePaint,Saharia22_Palette,zhange2023diffcollage,Ramesh21_Dalle,Saharia22_Imagen,Kawar22_JPEGCorrection,Nie22_AdvPurification}, and (3)~their relative ease of training~\citep{Dhariwal21_DiffusionBeatsGANs}. Given their success in generating images, there has been significant recent effort in developing graph denoising diffusion models (GDDMs) that accommodate discrete, structured graph data~\citep{Austin21_StructuredDenoiseing,digress_vignac22,Chen23_EDGE,Jo22_GDSS}, but there is limited work on supporting or even defining analogous graph-based editing tasks.

Inspired by the conceptual similarities between ``in-painting" (i.e., filling in masked regions of an image) and filling in missing nodes and edges conditioned on the observed graph, we observe that 
the distributional modeling capacity of a trained GDDM could be used to refine parts of an observed training graph that may be corrupted or incomplete. 
However, refining a \textit{single}
corrupted network raises additional challenges that preclude the applicability of existing GDDMs; these models assume that they are trained on a \textit{large set of clean graphs}, and have quadratic complexity over the number of nodes preventing them from scaling to more than a few hundred nodes~\citep{Jo22_GDSS, digress_vignac22}. 

\textbf{Our Work.} We focus on the challenging setting of training a GDDM on a single corrupted graph such that the trained model can then be used to perform various conditional graph editing tasks to refine the observed network. In particular, we propose a novel subgraph-based diffusion framework (\ours) that uses graph sampling, and global vs.\ local contextual information to support three editing tasks: \texttt{(T1)}~denoising, \texttt{(T2)}~expansion and \texttt{(T3)}~structured style-transfer tasks (Fig.~\ref{fig:overview}), which respectively remove extraneous edges, recover missing edges and regenerate graphs to support particular attributes. 
Our contributions are summarized as:
    
\noindent $\bullet$ \textbf{Single Graph Editing Tasks:}
We formally introduce three new graph editing tasks---denoising, expansion, style transfer---for single graph GDDMs (\S~\ref{sec:prob_formulation}). 

\noindent $\bullet$ \textbf{Subgraph Diffusion Framework:} We propose a novel, diffusion framework that utilizes graph sampling and global vs.\ local positional information to learn distributions over a single network and support the introduced editing tasks (\S~\ref{sec:approach}). 

\noindent $\bullet$ \textbf{Extensive Empirical Evaluation:} We introduce a diverse set of graph-specific metrics evaluating performance of the editing tasks, and extensively evaluate the ability of GDDMs and our proposed diffusion framework to perform them (\S~\ref{sec:exp}).

\section{Related Work \& Background} \label{sec:related-work}
For both image and graph generation tasks, diffusion models (DMs) have outperformed GAN~\citep{Cao18_MolGAN,Wang18_GraphGAN,Dhariwal21_DiffusionBeatsGANs,Liao19_GRAN} and autoencoding alternatives~\citep{Simonovsky18_GraphVAE,Zahirnia22_GraphVAEMM,You18_GraphRNN} in generating high fidelity, diverse inputs. Conceptually, DMs 
consist of two parts: (1) a \textit{forward} process which progressively adds noise to a sample and (2) a \textit{reverse} process which learns to remove noise from the corrupted sample. 
While several improvements have been suggested to \textit{vision} DMs to their improve trainability~\citep{Karras22_DesignSpaceOfDiffusion,Xiao22_GenerativeTrilemma}, fidelity and sampling speed~\citep{Rombach22_LatentDiffusion,Zhang23_FastSamplingExponential}, we provide a generic exposition that also applies to GDDMs here.  

Let $x^0 \sim q(x)$ be a sample from a data distribution. Then, we can define a forward process that gradually adds noise to $x^0$ over $T$ steps such that in limit $x^T$ can be seen as a sample from a predefined \textit{reference} distribution. Formally, let $q(x^t|x^{t-1})$ be one step of the forward process, then using the Markov property, we can condition on the data sample such that $q(x^1,\dots, x^T | x^0) = \prod_{t=1}^T q(x^t | x^{t-1})$. Image DMs~\citep{Dickstein15_Nonequilibrium} and early graph DMs~\citep{Niu20_EDPGNN,Jo22_GDSS} use an isotropic Gaussian for the reference distribution as it ensures that each step of the reverse process can be computed in closed form,  sampling $x^T$ does not require knowledge of $x^0$, and  the posterior has a closed form.

Then, the parameterized reverse process, $p_\theta$, learns to approximate $q(x^{t-1}|x^t)$ to iteratively convert a sample from reference distribution to a sample from $q(x)$. Since each step in the forward process is conditionally independent, we can compute the joint probability over the noisy latent $(x^0,\dots, x^T)$ as $p_\theta\left(x^{0:T}\right) =p\left(x^{T}\right)\prod_{t=1}^T p_\theta\left(x^{t-1}|x^{t}\right)$, where marginalizing out the trajectory models the data distribution, $p_\theta(x^{0})$. Thus, data can be generated by sampling from the reference distribution, and then using the learned reverse process at each step $T \rightarrow 0$. Several equivalent formulations have been proposed for training $p_\theta$ by minimizing the variational lower bound, which practically involve training a neural network to predict the noise at step $t$ \citep{Ho20_DDPM,Song19_EstimatingGradients,Song19_SlicedScoreMatching}. We refer the interested reader to a comprehensive survey by~\cite{Yang22_DiffusionSurvey}.

Since graphs are structured, discrete data, alternative reference distributions and corresponding forward processes have been proposed to preserve sparsity and discreteness. For example, DiGRESS~\citep{digress_vignac22} proposes a categorical distribution over node and edge types, while EDGE~\citep{Chen23_EDGE} uses an empty Erdos-Renyi graph. Both are more scalable than the Gaussian reference distribution used by GDSS~\citep{Jo22_GDSS}, and produce structured intermediate latent variables. We give more details for graph diffusion models in \Cref{app:expanded_related_work}. 

\section{Problem Formulation} \label{sec:prob_formulation}
In this section, we introduce the editing tasks that our framework supports (Fig.~\ref{fig:overview}). 

\textbf{Notations.} Let $\mathcal{G}(\mathcal{V},\mathcal{E},\mA,\mX)$ represent a graph where the tuple contains, respectively, the node set $\mathcal{V}$, edge set $\mathcal{E}$, adjacency matrix $\mA \in [0,1]^{|\mathcal{V}|}$, and node features $\mX \in \mathbb{R}^{|\mathcal{V}| \times d}$.  

\textbf{{Problem Statement.}} Let $\observedgraph(\observednode, \observededge)$ be an incomplete, noisy \textit{observed} network and $\targetgraph(\targetnode, \targetedge)$ be the corresponding, but unknown, complete, noise-free \textit{target} graph. Further, we assume that the majority of the observed graph is not corrupted or missing---i.e.,~{\small $|(\observededge \setminus \targetedge) \cup (\targetedge \setminus \observededge)| \le \min(\epsilon| \targetedge|,\epsilon|\observededge|)$}, where $\epsilon \ll 1$. Then, given $\observedgraph$, we seek to devise a scalable, unified GDDM framework that can support \texttt{(T1)}~expansion, \texttt{(T2)}~denoising, and \texttt{(T3)}~style transfer in order to recover the missing subgraphs and remove the extra subgraphs such that the refined graph better aligns with the unknown target graph $\targetgraph$. While these discrepancies could co-occur, we begin by considering them individually and assume all nodes are shared between the target and observed graphs. 

\textbf{\texttt{(T1)} Expansion:} Here, we assume that the observed graph is \textit{missing} some subgraph \textit{present} in the target graph. The task is to recover these missing edges given the observed graph. Formally, let $\missingedge = \targetedge \setminus \observededge$ correspond to the set of missing edges, where $0 < {|\missingedge|} < \epsilon$  and $|\observednode| = |\targetnode|$. 
Then, conditioning on $\observedgraph$, we wish to generate a graph $\gengraph$ that contains both $\observededge$ and $\missingedge$.

\textbf{\texttt{(T2)} Denoising:} 
Here, we assume that the observed graph has \textit{extra} edges present that are \textit{not} present in target graph. The task is to remove the additional edges given the observed graph. Formally, let $\additionaledge = \observededge \setminus \targetedge$ correspond to the set of additional edges, where $0 < \additionaledge < \epsilon$ and $|\observednode| = |\targetnode|$. Then, conditioning on $\observedgraph$, we wish to generate a graph $\gengraph$ that removes $\additionaledge$, but keeps the rest of the edges. 

\textbf{\texttt{(T3)} Style Transfer:} 
While expansion and denoising focus on editing $\observedgraph$ so that the generated graph better aligns with $\targetgraph$, style transfer is not strictly a refinement task. Instead, it focuses on adapting $\observedgraph$ to reflect a particular attribute (e.g. style) of interest. We include it in our editing framework since it provides a mechanism for making abstract conditional graph changes. 
{For example, we may be interested in how changing the number of triangles affects connectivity, but do not know apriori which particular edges are changed in response.} 
Formally, let $\texttt{Attr}$ be some attribute of interest and $\texttt{Attr}_D$ be the desired value. Then, given $\observedgraph(\observednode, \observededge)$ with observed value $\texttt{Attr}_O$, we wish to generate a graph, $\gengraph(\gennode, \genedge)$, such that $\gengraph$ obtains value $\texttt{Attr}_D$. We note that existing GDDMs cannot support this task.  

\section{Editing Networks with Subgraph Diffusion Models} \label{sec:approach}
Here, we describe our subgraph-based diffusion model (\ours) and editing framework (Sec. \ref{sec:editing}).

\subsection{\ours~ Framework}\label{sec:sgdm}

Our diffusion framework is motivated by the challenges that existing training diffusion models face in our problem setup (Sec.~\ref{sec:prob_formulation}). Specifically, the majority of image~\citep{Saharia22_Palette,Ramesh21_Dalle,Saharia22_Imagen} and graph diffusion models~\citep{digress_vignac22,Chen23_EDGE} are trained on very large datasets of generally clean samples to support diverse, high fidelity generation and conditional editing. However,
when training diffusion models are applied on a \textit{single} sample~\citep{Kulikov23_SinDDM,Nikankin23_sinfusion}, 
it is easier to overfit to the observed graph---i.e., the reverse process becomes effectively deterministic and only generates the observed graph. This is especially damaging as the observed graph is assumed to be corrupted or otherwise incomplete, and overfitting will lead the diffusion model to memorize these artifacts. Furthermore, while most single-\textit{image} diffusion models assume standard to moderately-sized training images, single-\textit{graph} diffusion models will likely need to accommodate very large training graphs. 
Lastly, since most parts of the graph are assumed to be complete and uncorrupted, it is unnecessary to regenerate the entire graph per edit.  

We propose Subgraph-based Diffusion Models (\ours) which rely upon \textit{subgraph sampling} and \textit{global context} to address these challenges, and flexibly support existing GDDMs in generating and editing large-scale networks (Algorithms~\ref{algo:sgdm_main}--\ref{algo:sgdm_editing}). Next, we discuss the key components of \ours.   

\begin{algorithm}[t]
\small
\caption{\ours: Subgraph-based Diffusion} 
\label{algo:sgdm_main}
\KwIn{$\observedgraph = (\mathcal{V},\mathcal{E},\mA, \mX )$, \texttt{Global-Context}, \texttt{Local-Context}, \texttt{Sampling}, $q$, $p_\theta$} 
\KwOut{$(q,p_\theta,\texttt{Hist-Global})$: trained subgraph diffusion model} 
$\mC \gets \texttt{Global-Context}(\observedgraph)$ \\
$\observedgraph \gets \mC$ \\
$\bm{\mathcal{S}} \gets \texttt{Sampling}(\observedgraph)$ \\
$\texttt{Hist-Global} = [0]^{|\observedgraph|}$ \\
$\texttt{Hist-GraphSize} = [0]^{|\text{Max Size}(|\bm{\mathcal{S}}|)}$ \\
\For{ $\mathcal{S}$ \KwTo $\bm{\mathcal{S}}$}{
    $\mQ \gets  \texttt{Local-Context}(\mathcal{S})$ \\
    $\mathcal{S} \gets \mQ$ \\
    $\texttt{Hist-Global}[\mC_{\mathcal{S}}] += 1$ \\
    $\texttt{Hist-GraphSize}[|\mathcal{S}|] += 1$ \\
    $\bm{\mathcal{S}} \gets (\mathcal{S} = \mA_\mathcal{S},\mX_\mathcal{S},\mC_\mathcal{S},\mQ) $ \\ 
}
$\texttt{Forward} \gets q\left(\mathcal{S}^1, \dots,\mathcal{S}^T\right) = \prod_{t=1}^T q\left(\mathcal{S}^t|\mathcal{S}^{t-1}\right)$ \\
$\texttt{Reverse} \gets p_\theta \left(\mathcal{S}^{0:T}\right) = p(\mathcal{S}^T)\prod_{t=1}^T p_\theta\left(\mathcal{S}^{t-1}|\mathcal{S}^{t},\mC_{\mathcal{S}^0}\right)$ \\
$\min_\theta \sum_{\mathcal{S} \in \bm{\mathcal{S}}}\mathbb{E}_{q(\mathcal{S}_0)q(\mathcal{S}_{1:T}|\mathcal{S}_0)} \left[-\log\frac{p_\theta(\mathcal{S}_{0:T})}{q(\mathcal{S}_{1:T}|\mathcal{S}_0)} \right]$ \\ 
\Return $(q, p_\theta,\texttt{Hist-Global},\texttt{Hist-GraphSize})$
\end{algorithm}

\begin{algorithm}[t]
\small
\caption{Editing with \ours} 
\KwIn{$\mathcal{S}_{O}=\left(\observednode, \observededge, \mA_{O}, \mX_{O}, \mC_{O} \right)$: subgraph that will be edited, $f_\psi:(\mathcal{S},t) \rightarrow y$: trained attribute regressor, $\lambda$: guidance strength} 
\KwOut{$\mathcal{S}_{G}=\left(\mA_{G}, \mX_{G}, \mC_{O} \right)$: edited graph.} 
   $n \gets |\mathcal{S}_{O}|$ \\
   $\mathcal{S}^{t=T} \gets q_n(\cdot)$ \\
    \For{t = T \KwTo 0}{
        $\mQ^t \gets \texttt{Local-Context}(\mathcal{S}^t)$ \\
        $\mathcal{S}^t \gets \mQ^t $ \\ 
        $\mathcal{S}^{t-1} \gets p_\theta(\mathcal{S}^t,\mC)$ \\
        $ (\mA^{t-1}, \mX^{t-1},\mC) \gets \mathcal{S}^{t-1}$ \\
        \Comment{Expansion.} 
        $\mA^{t-1} \gets (1 - \mA) \odot \mA^{t-1} + \mA$ \\
        $\mX^{t-1} \gets (1 - \mX) \odot \mX^{t-1} + \mX$ \\
        \Comment{Denoising.} 
        $\mA^{t-1} \gets \mA \odot \mA^{t-1} $ \\
        \Comment{Style Transfer. \citep{digress_vignac22,Dhariwal21_DiffusionBeatsGANs}}  
        $\widetilde{y} = f(\mathcal{S}^t,t)$ \\
        $p_\psi(\widetilde{y}|\mathcal{S}^{t-1}) \propto
        exp(-\lambda \langle \nabla \mathcal{S}^t || \widetilde{y} - y||^2, \mathcal{S}^{t-1}\rangle)$ \\
        $\mathcal{S}^{t-1} \sim p_\theta(\mathcal{S}^{t-1}|\mathcal{S}^t) p_\psi(\widetilde{y}|\mathcal{S}^{t-1})$ 
    }
\Return $\mathcal{S}_G = (\mA^{t},\mX^{t},\mC_O)$ 
\end{algorithm}\label{algo:sgdm_editing}

\textbf{Subgraph Sampling.} Existing GDDMs have demonstrated impressive performance in modeling large distributions of \textit{small} graphs but are limited by the quadratic memory and runtime complexity incurred by considering every potential edge during generation. This is further exacerbated when using expressive but memory-intensive graph transformer architectures during training. While there has been some recent work on improving the scalability of GDDMs by only taking the gradient with respect to a subset of active nodes~\citep{Chen23_EDGE}, such approaches are nonetheless destined to run out-of-memory on production-scale networks as even a subset of active nodes will be too large to compute the backward pass. Moreover, as more sophisticated GDDMs are proposed, it is beneficial to support these methods on large-scale networks as well. 

To this end, we use subgraph sampling and stitching to convert the task of learning on a single large-scale network to learning a distribution over a collection of subgraphs. 
Namely, given the observed graph, $\observedgraph$, we use a subgraph sampling function: $\texttt{SAMPLE}:\observedgraph \rightarrow \bm{\mathcal{S}}$, to return a set of subgraphs, $\bm{\mathcal{S}}$, such that 
(1)~the average size of a subgraph $\mathcal{S}$ is much smaller than $\observedgraph$,  i.e., $\left(\frac{1}{|\bm{\mathcal{S}}|}\sum_{\mathcal{S}\in\bm{\mathcal{S}}} |\mathcal{S}|\right) \lll |\observedgraph|$;
(2)~each node is represented \textit{at least} once through sampling---$\forall v \in \observednode, |\{\mathcal{S} | v \in (\mathcal{S} \ni \bm{\mathcal{S}}\}| > 1$; and (3)~for trainability, $\mathcal{S}$ is connected.  Graph partitioning and clustering methods, as well as ego-network sampling are candidates for sampling, potentially followed by post-processing to ensure the connectivity requirement. 
In case the obtained subgraphs remain prohibitively large, an
optional subsampling function can be applied to reduce the size of $\mathcal{S}$ to a predefined maximum size $N_{max}$. Formally, define $\texttt{SUBSAMPLE}: \mathcal{S} \rightarrow \widetilde{\mathcal{S}}$, such that $|\widetilde{\mathcal{S}}| \le N_{max}$ and 
$\widetilde{\mathcal{S}}$ remains connected. 

Given the subgraphs, we can use any existing GDDM to define $p_\theta$, $q$, and the reference distribution~\citep{digress_vignac22,Chen23_EDGE,Jo22_GDSS} over $\mathcal{S}$ as follows, including those that may have been previously prohibited through the large memory complexity: 
\vspace{-5pt}
{\small 
\begin{align}
&\text{Forward Process: } q\left(\mathcal{S}^t_k(v)|\mathcal{S}^{t-1}_k(v)\right) \label{eq:sgdm1}\\
&\text{Reverse Process: }  p_\theta\left(\mathcal{S}^{t-1}_k(v)|\mathcal{S}^{t}_k(v)\right)\label{eq:sgdm2}\\
&\text{Joint Prob: } \textstyle  p\left(\mathcal{S}^{T}_k(v)\right)\prod_{t=1}^T p_\theta\left(\mathcal{S}^{t-1}_k(v)|\mathcal{S}^{t}_k(v)\right).\label{eq:sgdm3}
\end{align}
}
We note that while our primary objective is \textit{not} to learn a large-network generative model, \ours can be used for this purpose as it has been shown that it is possible to reconstruct a graph from a reasonably-sized collection of its random subgraphs with sufficient coverage~\citep{Manvel76_ReconstructingSubgraphs, McGregor22_RandomSubgraphRecon}. We discuss ``stitching'' at the end of the subsection. 

\textbf{Context.} While Equations~(\ref{eq:sgdm1},\ref{eq:sgdm2},\ref{eq:sgdm3}) define a complete graph diffusion model over $\bm{\mathcal{S}}$, we can further improve its generative capabilities by introducing information that can differentiate between similar subgraphs. 
It is possible that there exist subgraphs $\mathcal{S}_i, \mathcal{S}_j \in \bm{\mathcal{S}}$ which are structurally similar but located in different parts of the network.
To handle such cases effectively, we design 
\ours to leverage both \textit{global} context (e.g., how close $\mathcal{S}_i$ is to $\mathcal{S}_j$ in $\observedgraph$, what $\mathcal{S}_i$'s surroundings are) and 
\textit{local} context (i.e., the structural characteristics of $\mathcal{S}_i$) for editing tasks. 
For example, when editing a corrupted subgraph, it may be more helpful to leverage information from a closer, structurally similar subgraph than one that is further away; global context helps us capture this and ensures the edited graph remains consistent with respect to its surroundings. 

To improve model performance and effectively provide $\texttt{Local-Context}:\mathcal{S} \rightarrow \mathbb{R}^{|\mathcal{S}| \times d}$, existing GDDMs~\citep{digress_vignac22,Chen23_EDGE} use spectral and structural features (e.g., cycle counts, number of triangles, Laplacian vectors, degree distributions). 
For global context, let $\texttt{Global-Context}:\observedgraph \rightarrow \mathbb{R}^{|\observedgraph| \times d}$ be a function that computes a global-context vector ($\mC_v$) for each node $v\in\observednode$ such that (i) $\forall{u,v,q u \neq v \in \observednode}, \mC_u \neq \mC_v$, and (ii) for most $u,v, (u \neq v) \in \observednode$, $|\mC_u$ - $\mC_v| \le |\mC_v$ - $\mC_q|$ if $d(u,v) \le d(v,q)$ for some graph distance $d$. 
To incorporate global information, we modify $p_\theta$ to take $\mC_{\mathcal{S}}$ as an additional feature (as described in the appendix). Notably, when generating new samples, we first sample  $n$ subgraph global-context vectors from the empirical distribution of global-context vectors in $\bm{\mathcal{S}}$, and then proceed with the reverse process. This allows us to control the broader context in which graphs are generated. (\Cref{algo:sgdm_main}).

\textbf{Stitching.}\label{sec:stitching} 
As we mentioned above, it is possible to reconstruct graphs given a collection of random subgraphs \citep{Manvel76_ReconstructingSubgraphs,Gupta12_ReconConjecture,McGregor22_RandomSubgraphRecon} with enough samples and provided there is a mechanism for identifying nodes across the subgraphs so that edges may be coalesced. Since we required \texttt{Global-Context} to be unique $\forall v \in \observednode$, we can use the global-context vectors to perform this identification and then define a \texttt{COALESCE}: $\widetilde{\bm{\mathcal{S}}}\rightarrow \widetilde {\observedgraph}$ function accordingly, e.g., take the union over collection of edges. (\Cref{algo:sgdm_stitching}). 

\subsection{Supported Editing Tasks}
\label{sec:editing}
We now discuss how to perform the editing tasks introduced in Sec.~\ref{sec:prob_formulation} for both \ours and single-graph diffusion models. We assume that we have a pretrained diffusion model, $p_\theta$, and the portion of the graph that needs to be edited is provided, $\mathcal{S}(\mathcal{V}, \mathcal{E}, \mA, \mX, \mC)$. 

\texttt{(T1)} \textbf{Expansion with \ours.} For expansion, the reverse process needs to recover missing edges or subgraphs, given the specified, observed subgraph. 
Conceptually, this task is similar to ``in-painting" where a diffusion model is used to fill-in masked regions of an image so that the resulting image is consistent with the surrounding context. 
Our approach is inspired by a computer vision approach~\citep{Lugmayr22_RePaint} that uses the DM to only fill in the masked regions of the images; 
it only modifies the reverse process and ensures consistency with the starting sample by incorporating information from the unmasked regions throughout the reverse process. 
For graphs, the masked region is defined by the connectivity and node features of the provided, starting subgraph, $\mathcal{S}(\mA, \mX, \mQ)$, and unmasked region is implicitly defined by the potential $|\mathcal{S}|^2$ - $\mA$ edges of $\mathcal{S}$. Formally, for each step $t$ in the reverse process,  
$\mA^{t-1} \gets (1 - \mA) \odot \mA^{t-1} + \mA,$
where the full process is shown in Algorithm \ref{algo:sgdm_editing}.

\texttt{(T2)} \textbf{Denoising with \ours.} 
While expansion assumes that all the existing edges  $\mathcal{S}(\mA, \mX, \mC)$ are reliable, and can be used to guide the reverse process, this is not the case when performing denoising. Indeed, when performing denoising, it is unknown \textit{which} subset of $\mA$ can be considered reliable. This makes it challenging to ensure that the generated sample is consistent with the starting sample. On the other hand, while expansion requires correctly identifying which of the possible $|\mathcal{S}|^2$ - $\mA$ edges should be added to $\mathcal{S}$, denoising requires only identifying which of the $\mA$ edges are in fact correct. Since graphs are often sparse, this entails that the possible ``solution" space when denoising is significantly smaller than that when performing expansion. 
We propose to use an in-painting style approach, but instead of filling-in regions given $\mathcal{S}$, we attempt to remove edges that do not belong: $\mA^{t-1} \gets \mA \odot \mA^{t-1}$. 
This ensures that the sample generated at the end of the reverse process will only contain edges that belong to $\mathcal{S}$, and, due to the impressive generative capacity of the pretrained diffusion model, resulting sample may also remove the extraneous edges. 

\texttt{(T3)} \textbf{Style Transfer with \ours.}  
Consider an example where we are interested in understanding how changing a node's attribute will lead to changes in its connectivity. With an existing single large-graph GDDM, the entire graph would be regenerated in an attempt to reflect these changes despite the expectation that edits will be relatively local~\citep{Newman10_NetworksBook}. In contrast, with \ours, we can provide the relevant subgraph (i.e., k-hop egonet) of the specified node and regenerate to reflect the changed attribute. This is particularly important for scalability as we do not consider an in-painting approach here, but instead leverage classifier guidance~\citep{Dhariwal21_DiffusionBeatsGANs, Kawar23_RobustClassifierGuidance, digress_vignac22} which requires taking the gradient of the generated graph while attempting to minimize the loss of a time-dependent classifier.  
We build on a conditional generation pipeline~\citep{digress_vignac22} that is unable to perform style transfer on large networks due to memory issues during training. To overcome these challenges, in brief, we first train a time-dependent regressor, $f_\psi(\mathcal{S},t)$ to predict the attribute of interest, then we use the regressor loss, $(p_\psi(\widetilde{y}|\mathcal{S}^{t-1}))$, to guide the reverse process 
$\mathcal{S}^{t-1} \sim p_\theta(\mathcal{S}^{t-1}|\mathcal{S}^t) p_\psi(\widetilde{y}|\mathcal{S}^{t-1})$  (\Cref{algo:sgdm_editing}).

\section{Experiments}\label{sec:exp}
In this section, we evaluate the ability of graph diffusion models to perform editing tasks, and demonstrate the benefits of using \ours. Specifically, we seek to answer the following research questions: 
\textbf{(RQ1)}~Is there a benefit to \texttt{SAMPLING} on the expansion and denoising tasks?
\textbf{(RQ2)}~Is there a benefit to using \texttt{GLOBAL-CONTEXT}  in our proposed graph editing tasks? 
\textbf{(RQ3)}~How do different GDDM backbones affect the performance of \ours? 
Below, we first describe our evaluation setup.

\subsection{Evaluating Refinement Tasks}
\label{sec:eval_metrics}

To the best our knowledge, we are the first to consider using graph editing to perform refinement on a single, partially observed network. 
Therefore, we begin by introducing the following new metrics.

\subsubsection{Novel Evaluation Metrics}
Evaluating the ability of generative graph models to recover $\targetgraph(\targetnode, \targetedge)$ through conditional editing tasks presents several unique difficulties in comparison to similar editing tasks in computer vision. While repainted images are assessed for ``realism", which considers how plausible an in-painting solution is given the unmasked region, and ``diversity", which considers the perceptual dissimilarity of generated solutions, it is non-trivial to define such quantities for edited graphs. Indeed, in the case of a subgraph missing a single edge, generating a diverse set of alternative subgraphs may introduce unnecessary changes that do not align with $\targetgraph$. Furthermore, it is not possible to perform user-studies or visual inspection to determine if the edited graph maps to realistic solutions due to their size and the abstract nature of structured data. To this end, we introduce the following novel metrics for assessing the quality of expanded or denoised graphs. We discuss style transfer evaluation separately.
\begin{table}[t]
{\scriptsize
\centering
\caption{\label{table:graph_stats} \textbf{Dataset Statistics}}
\vspace{-5pt}
\resizebox{\columnwidth}{!}{\begin{tabular}{l ccc}
\toprule
\textbf{Dataset} & \textbf{\# Graphs} & \textbf{\# Nodes} & \textbf{\# Edges} 
\\
\midrule
BA-Shapes & 1 & 700 & 3,944  \\
PolBlogs & 1 & 1,222 & 33,431  \\
CORA & 1 & 2,485 & 10,138 \\
\bottomrule
\end{tabular}%
}
}
\end{table}

$\bullet$ \textbf{Consensus:} If the score function has learned to model the true data-generating process underlying the unobserved target graph, then it is expected that the majority of generated samples will concur that the missing edges should be generated. A higher consensus score, $\frac{1}{|\missingedge|}\sum_{[i,j] \in \missingedge}\frac{1}{R}\sum_{r=1}^R \mathds{1}\left[{\genedge}^r_{[i,j]} = {\targetedge}_{[i,j]}\right]$, indicates that the majority of samples are able to generate the particular set of missing edges, as desired.

$\bullet$ \textbf{Edge Overlap:} While \ours~allows us to generate aligned graphs by conditioning with the positional information, we note that other methods may not produced such graphs. Therefore, after performing some alignment, we measure how much of the generated graph overlaps with the conditioning graph. Since, measuring overlap is an $NP$-hard problem, we approximate it as $\frac{1}{R}\sum_{r=1}^R \mathds{1}\left[{\genedge}^r\odot {\observededge}\right]$, using the binary adjacency matrix representation of the graph.

$\bullet$ \textbf{Diversity:} While producing a diverse set of generated solutions may not be as high priority in graph editing tasks as in image editing tasks, it is still useful to measure to understand if the generative model has converged to a single solution or collapsed to single model: $\frac{1}{R(R-1)}\sum_{r_1=1}^R \sum_{r_2=1}^R \mathds{1}\left[{\genedge}^{r_1} \neq {\genedge}^{r_2} \right]$.

$\bullet$ \textbf{Sparsity:} Given that we seek to recover missing edges conditioned on the observed graph, it is possible to trivially recover all missing edges if the generated samples are fully connected. Thus, we measure the fraction of extra edges generated with respect to the target graph: $\frac{1}{R}\sum_{r=1}^R\frac{|\genedge^r|}{|\targetedge|}$. 

While the desired sparsity score is closer to 1, the range of sparsity is $[\frac{|\observededge|}{|\targetedge|}, \frac{|\observednode|^2}{|\targetedge|}]$, corresponding to generated samples which do not add any edges given the conditioning graph and those that generate all possible edges.

\subsubsection{Experimental Setup}
\noindent \textbf{{Datasets.}} 
For the editing tasks, we consider 3 large, single networks---BA-Shapes, PolBlogs and CORA (\Cref{table:graph_stats})---and corrupt them to create the incomplete, noisy observed graphs. Namely, we randomly add or remove 10\% of edges in the CORA or Pol-Blogs dataset, while enforcing that the graph remains connected. In order to assess performance on a structured corruption process, we also evaluate the BA-Shapes dataset \citep{Ying19_GNNExplainer}, which consists of 80 house-motifs attached to a 300-node Barbasi Albert random graph. Here, we corrupt 5\% of the edges in the house-motifs. Since the house-motifs have a fixed structure, we can better understand if the diffusion model is able to capture the underlying data-generating process in order to refine the graph. Given the set of the removed or added edges, we are able to compute the metrics introduced above (Sec. \ref{sec:eval_metrics}).

{\textbf{Baselines.}}  Given that editing a single network is a relatively new problem, there are no direct baselines, so we adapt existing graph diffusion models to the best of our abilities. EDGE~\citep{Chen23_EDGE}, which conditions the reverse process using the empirical degree distribution is our most direct comparison and, as we discuss below, helps us show the benefits of \ours. 

{\textbf{\ours~ Variants.}}
While \ours~was presented as a general framework in Sec. \ref{sec:sgdm}, 
we evaluate specific variants that are
well-suited for our editing tasks. We define \texttt{SAMPLING} using node-centric 2-hop ego-networks; 
\texttt{SUBSAMPLING} by (1)~randomly sampling 200 nodes per subgraph, (2)~inducing the corresponding subgraph, and (3)~taking the largest connected component; 
and \texttt{GLOBAL-Context} by computing the top-two Laplacian positional encodings (LPEs) \citep{Dwivedi20_BenchmarkingGNNs,Dwivedi22_LearnablePE,Wang22_ESLPE} on $\observedgraph$. We select ego-networks for \texttt{SAMPLING} as it aligns well with use cases where particular nodes are untrustworthy or corrupted.  
Our \texttt{SUBSAMPLING} strategy is selected for ease of use and efficiency. We chose LPEs for global context as they have been shown to be theoretically expressive in distinguishing between nodes. We consider 3 different backbone GDDMs (DiGRESS~\citep{digress_vignac22}, EDGE~\citep{Chen23_EDGE} and GDSS~\citep{Jo22_GDSS}) in our evaluation to demonstrate the flexibility the proposed framework. \texttt{Local-Context} is defined according to backbone GDDM. We discuss more variants in \Cref{app:sgdm_details}.

\subsubsection{Results}
Here, we answer the aforementioned research questions. Scores closer to ``1" indicate better performance.  

\mypar{\textbf{(RQ1) Benefits of Sampling.}} In Table \ref{tab:benefits_of_sampling}, EDGE is compared against \ours~when using a EDGE backbone and NO \texttt{GLOBAL-CONTEXT} to determine if learning over a collection of subgraphs is more effective than directly learning on a single graph. We observe that \ours significantly outperforms EDGE on Diversity, Sparsity and Edge Overlap on the denoising task. While EDGE achieves better Consensus, it does so by removing too many edges. On expansion, both struggle to perform well. We posit that this is a side-effect of guiding the reverse process using a degree distribution, as other GDDM backbones perform better.

\mypar{\textbf{(RQ2) Benefits of Global-Context.}} 
In Table~\ref{tab:benefits_of_context}, we see that \ours~(with \texttt{GLOBAL-CONTEXT}) continues to outperform EDGE with or without \texttt{GLOBAL-CONTEXT}. Moreover, compared against \ours~(without \texttt{GLOBAL-CONTEXT}) in Table~\ref{tab:benefits_of_sampling}, we see that including context improves Sparsity and Edge-Overlap on both Cora and PolBlogs. Recall  that context also enables effective stitching for large network generation, though it is not the main focus of this work.

\begin{figure}[t]
    \centering
    \includegraphics[width=0.99\columnwidth]{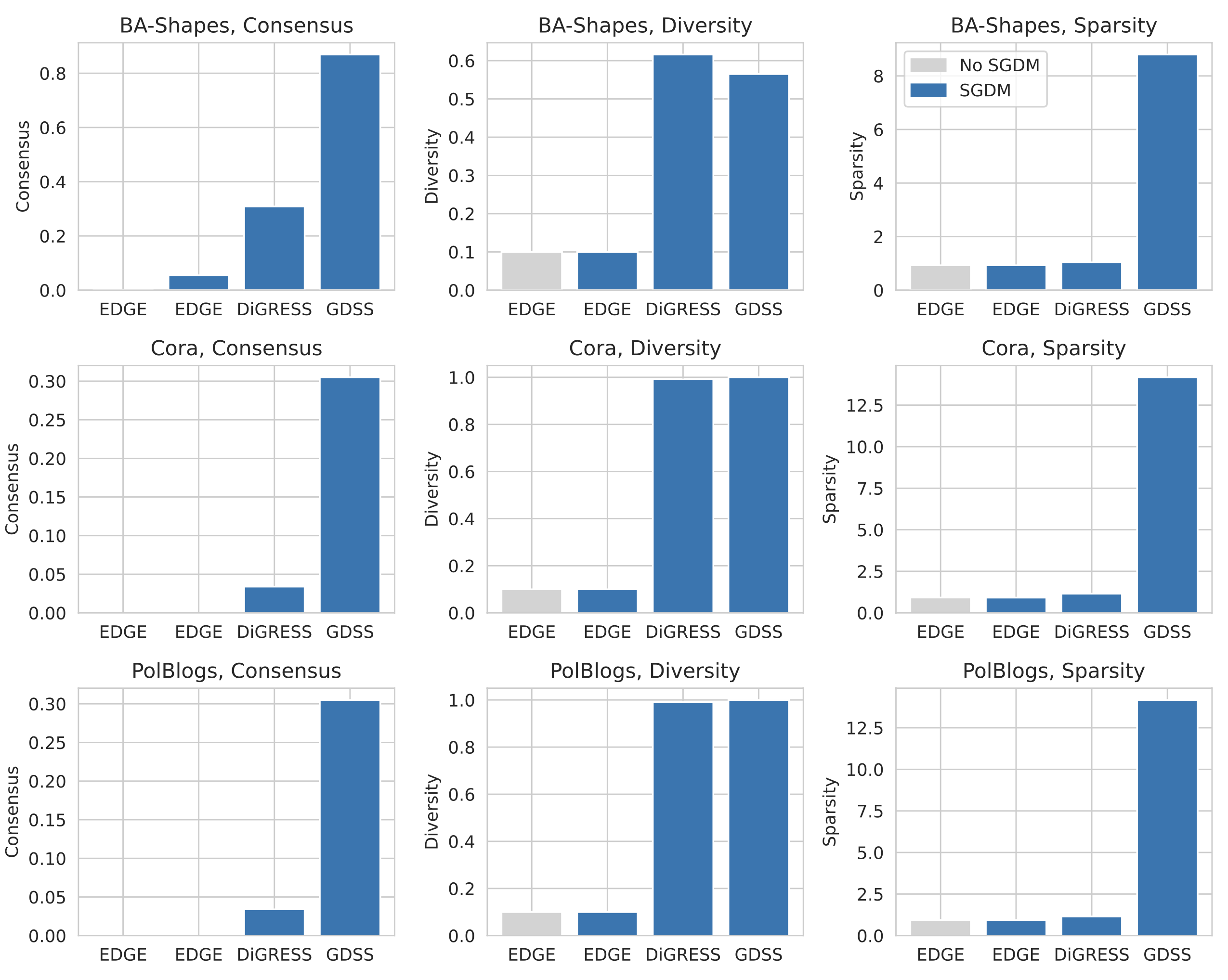}
    \caption{\textbf{Choice of Diffusion Models.} \ours continues to outperform EDGE  when using GDSS and DiGRESS as backbones. Indeed, these backbones improve the Consensus and Diversity as well.}
    \label{fig:expansion_choice_diffusion_models}
\end{figure}

\mypar{\textbf{(RQ3) Effect of Backbone.}} To demonstrate the flexibility of our proposed framework, we evaluate the performance of \ours on expansion when using GDSS and DiGRESS backbones (see \ref{app:extra_experiments} for results on denoising). In Fig.~\ref{fig:expansion_choice_diffusion_models}, we see that both DiGRESS and GDSS achieve better Consensus and Diversity than EDGE. While GDSS has significantly greater Sparsity than other methods, this is not unexpected as GDSS treats the adjacency matrix as a continuous variable, leading to a densely generated graph. Nonetheless, \ours~continues to outperform EDGE, the only comparable single-graph method.

\begin{table}[t]
\centering
\caption{\textbf{Benefits of \ours.} Compared against EDGE, we see that using \ours+EDGE (gray rows) improves denoising performance. EDGE uses the degree distribution to guide the reverse process. This has negative effects on \texttt{(T1)} expansion, as it constrains the model to satisfy an expected number of edges. Using other DM backbones helps resolve this problem.}
\label{tab:benefits_of_sampling}
\resizebox{\columnwidth}{!}{%
\begin{tabular}{llHccHHcccc}
\toprule
\textbf{Dataset} &
  \textbf{Task} &
  \textbf{Method} &
  \textbf{SGDM} &
  \textbf{Global-Ctx} &
  \textbf{Alignment} &
  \textbf{Conditioning} &
  \multicolumn{1}{l}{\bf Consensus} &
  \multicolumn{1}{l}{\bf Diversity} &
  \multicolumn{1}{l}{\bf Sparsity} &
  \multicolumn{1}{l}{\bf Edge Overlap} \\
\midrule
BA-Shapes & Denoising & EDGE & \xmark  & \xmark  & \xmark  & Obs-Subst & \textbf{0.954} & 0.834 & 0.055  & 0.0427 \\
\rowcolor{Gray} BA-Shapes & Denoising & EDGE & \checkmark & \xmark  & \xmark  & Obs-Subst & 0.751 &  \textbf{0.992} & \textbf{0.364} & \textbf{0.3427} \\ 
\midrule
BA-Shapes & Expansion & EDGE & \xmark & \xmark  & \xmark  & Obs-Subst & 0.000 & 0.100 & 0.930 & 1.000 \\ 
\rowcolor{Gray} BA-Shapes & Expansion & EDGE & $\checkmark$ & \xmark  & \xmark  & Obs-Subst & \textbf{0.055} & 0.100  & 0.930   & 1.000       \\
\midrule
Cora      & Denoising & EDGE & \xmark  & \xmark  & \xmark  & Obs-Subst & \textbf{0.916} & 0.915 & 0.115  & 0.0957  \\
 \rowcolor{Gray} Cora      & Denoising & EDGE & $\checkmark$ & \xmark  & $\checkmark$ & Obs-Subst & 0.587 & \textbf{1.000} & \textbf{0.591}  & \textbf{0.5851} \\
\midrule
Cora      & Expansion & EDGE & \xmark  & \xmark  & \xmark  & Obs-Subst & 0.000     & 0.100   & 0.922  & 1.000       \\
\rowcolor{Gray} Cora      & Expansion & EDGE & $\checkmark$ & \xmark  & \xmark  & Obs-Subst & 0.000     & 0.100   & 0.925  & 1.000       \\
\midrule
PolBlogs  & Denoising & EDGE & \xmark  & \xmark  & \xmark  & Obs-Subst & \textbf{0.992} & 0.702 & 0.0411 & 0.0412  \\
\rowcolor{Gray} PolBlogs  & Denoising & EDGE & $\checkmark$ & \xmark  & \xmark  & Obs-Subst & 0.713 & \textbf{1.000}     & \textbf{0.340}   & \textbf{0.3156} \\
\midrule
PolBlogs  & Expansion & EDGE & \xmark  & \xmark  & \xmark  & Obs-Subst & 0.000     & 0.100   & 0.941  & 1.000       \\
\rowcolor{Gray} PolBlogs  & Expansion & EDGE & $\checkmark$ & \xmark  & \xmark  & Obs-Subst & 0.000     & 0.100   & 0.939  & 1.000       \\
\bottomrule
\end{tabular}%
}
\end{table}

\begin{table}[t]
\centering
\caption{\textbf{Benefits of \texttt{Global-Context}.} We see that \ours with \texttt{Global-Context} (gray rows) continues to outperform single-graph EDGE with or without additional context. Recall that context is also useful for stitching to generate large graphs.}
\label{tab:benefits_of_context}
\resizebox{\columnwidth}{!}{%
\begin{tabular}{llHccHHcccc}
\toprule
\textbf{Dataset} &
  \textbf{Task} &
  \textbf{Method} &
  \textbf{SGDM} &
  \textbf{Global-Ctx} &
  \textbf{Alignment} &
  \textbf{Conditioning} &
  \multicolumn{1}{l}{\bf Consensus} &
  \multicolumn{1}{l}{\bf Diversity} &
  \multicolumn{1}{l}{\bf Sparsity} &
  \multicolumn{1}{l}{\bf Edge Overlap} \\
\midrule
BA-Shapes & Denoising & EDGE & \xmark  & \xmark  & \xmark  & Obs-Subst & 0.954 & 0.834 & 0.055  & 0.0427 \\
BA-Shapes & Denoising & EDGE & \xmark  & $\checkmark$ & $\checkmark$ & Obs-Subst & \textbf{0.996} & 0.733 & 0.019  & 0.0152 \\
\rowcolor{Gray}BA-Shapes & Denoising & EDGE & $\checkmark$ & $\checkmark$ & \xmark  & Obs-Subst & 0.341 & \textbf{0.928} & \textbf{0.762}  & \textbf{0.7210}   \\
\midrule
BA-Shapes & Expansion & EDGE & \xmark & \xmark  & \xmark  & Obs-Subst & 0.000 & 0.100 & 0.930 & 1.000 \\ 
BA-Shapes & Expansion & EDGE & \xmark  & $\checkmark$ & \xmark  & Obs-Subst & 0.000 & 0.100 & 0.930 & 1.000 \\
\rowcolor{Gray}BA-Shapes & Expansion & EDGE & $\checkmark$ & $\checkmark$ & \xmark  & Obs-Subst & \textbf{0.055} & 0.100   & 0.930   & 1.000       \\
\midrule
Cora & Denoising & EDGE & \xmark  & \xmark  & \xmark  & Obs-Subst & 0.916 & 0.915 & 0.115  & 0.0957  \\
Cora      & Denoising & EDGE & \xmark  & $\checkmark$ & \xmark  & Obs-Subst & \textbf{0.965} & 0.798 & 0.057  & 0.0486 \\
\rowcolor{Gray}Cora      & Denoising & EDGE & $\checkmark$ & $\checkmark$ & \xmark  & Obs-Subst & 0.754 & \textbf{1.000}     & \textbf{0.442}  & \textbf{0.4376} \\
\midrule
Cora      & Expansion & EDGE & \xmark  & \xmark  & \xmark  & Obs-Subst & 0.000     & 0.100   & 0.922  & 1.000       \\
Cora      & Expansion & EDGE & \xmark  & $\checkmark$ & \xmark  & Obs-Subst & 0.000     & 0.100   & 0.922  & 1.000       \\
\rowcolor{Gray}Cora      & Expansion & EDGE & $\checkmark$ & $\checkmark$ & \xmark  & Obs-Subst & 0.000     & 0.100   & 0.918  & 1.000       \\
\midrule
PolBlogs  & Denoising & EDGE & \xmark  & \xmark  & \xmark  & Obs-Subst & \textbf{0.992} & 0.702 & 0.0411 & 0.0412  \\
PolBlogs  & Denoising & EDGE & \xmark  & $\checkmark$ & \xmark  & Obs-Subst & \textbf{0.993} & 0.598 & 0.023  & 0.0218 \\
\rowcolor{Gray}PolBlogs  & Denoising & EDGE & $\checkmark$ & $\checkmark$ & \xmark  & Obs-Subst & 0.375 & \textbf{1.000}  & \textbf{0.640} & \textbf{0.5927} \\
\midrule
PolBlogs  & Expansion & EDGE & \xmark  & \xmark  & \xmark  & Obs-Subst & 0.000     & 0.100   & 0.941  & 1.000       \\
PolBlogs  & Expansion & EDGE & \xmark  & $\checkmark$ & \xmark  & Obs-Subst & 0.000     & 0.100   & 0.938  & 1.000       \\
\rowcolor{Gray}PolBlogs  & Expansion & EDGE & $\checkmark$ & $\checkmark$ & \xmark  & Obs-Subst & 0.000     & 0.100   & 0.939  & 1.000  \\
\bottomrule
\end{tabular}%
}
\end{table}

\subsection{Graph Style Transfer Results}\label{sec:exp-results-style-transfer}
Unlike \texttt{(T1)} expansion and \texttt{(T2)} denoising, which seek to use the generative model to refine subgraphs so that they may better align with \targetgraph, \texttt{(T3)} Style Transfer instead modifies subgraphs so that they may align with some specified property of interest. Thus, we introduce a task-specific experimental and evaluation setup. 

\textbf{Style Transfer Tasks. } We consider three different graph properties to modify through style-transfer: (1) Sum of Degrees: Given a subgraph with total degree $D_T$, we wish to modify it so that the resultant graph has $D_T + 3$ as the total degree, (2) Maximum Degree: Given a subgraph with maximum degree $D_M$, we wish to modify it so that the result graph has $D_M + 3$ as the maximum degree, and (3) Number of Triangles:  Given a subgraph with triangle count $\tau_c$, we wish to modify it so that the result graph has $\tau_c + 3$ as the total number of triangles. These properties are selected because they can correspond to abstract concepts that may be relevant to downstream tasks. For example, in a social network, changing the sum of degrees or maximum degree may simulate a new user becoming more connected over time, while changing the number of triangles can reflect how friends-of-friends may become connected.  

\textbf{Experimental Setup.} For all style transfer tasks, we first train a regressor, $p(y|G_t,t)$ that predicts the target (e.g., max degree) from the graph at time $t$, and use it with $\lambda=100$ to guide the reverse process as shown in \Cref{algo:sgdm_editing}. Given that predicting graph properties using GNNs can be difficult, we report results for three different regressor backbones---GCN, GIN, and Graph Transformer. 
Furthermore, we consider generative models that have been trained with and without positional encodings. We emphasize that while our setup does follow that of \citeauthor{digress_vignac22} and \citeauthor{Dhariwal21_DiffusionBeatsGANs}, without the SGDM framework this task is meaningless on single  networks, as localized edits cannot be performed. 
\textbf{Evaluation.} In Table~\ref{table:graph-styletransfer-results-bashapes}, we report the average of the mean absolute error between the target property of interest and the resulting property from the generated graph for 100 samples. Here we show results on the BAShapes dataset when DiGRESS is used as the underlying diffusion model, and provide additional results including qualitative examples in the \ref{app:extra_experiments}.
\begin{table}[h]
\centering
\caption{\textbf{Graph Style Transfer.} While \ours can support localized style transfer, single-graph DMs like EDGE cannot. Moreover, we find that incorporating \texttt{Global-Context} can helps improve performance on two out of three style transfer tasks.}
\label{table:graph-styletransfer-results-bashapes}
\vspace{-2mm}
\footnotesize
\resizebox{\columnwidth}{!}{\begin{tabular}{lHcHccH}
\toprule
\textbf{Task} & \textbf{Dataset} & \textbf{\texttt{Global-Context?}} & \textbf{Conditioning?} & \textbf{Regressor Type} & \textbf{MAE}$(\downarrow)$ & \textbf{Lamda} \\ \midrule
\multirow{6}{*}{Sum-Degree} & BAShapes & \xmark & \xmark & GCN    & 6.46  \\
                            & BAShapes & \xmark & \xmark & GIN    & 6.81  \\
                            & BAShapes & \xmark & \xmark & GTrans & 6.48  \\
                            \cmidrule{2-7}
                            & BAShapes & \checkmark  & \xmark & GCN    & 3.78  \\
                            & BAShapes & \checkmark  & \xmark & GIN    & 3.62  \\
                            & BAShapes & \checkmark  & \xmark & GTrans & $\bm{3.45}$  \\
\midrule
\multirow{6}{*}{Max-Degree} & BAShapes & \xmark & \xmark & GCN    & 24.73 \\
                            & BAShapes & \xmark & \xmark & GIN    & 45.09 \\
                            & BAShapes & \xmark & \xmark & GTrans & 24.63 \\
                            \cmidrule{2-7}
                            & BAShapes & \checkmark  & \xmark & GCN    & $\bm{9.3}$   \\
                            & BAShapes & \checkmark  & \xmark & GIN    & 11.04 \\
                            & BAShapes & \checkmark  & \xmark & GTrans & 11.44 \\
\midrule
\multirow{6}{*}{Num. Triangles}  & BAShapes & \xmark & \xmark & GCN    & $\bm{9.18}$  \\
                            & BAShapes & \xmark & \xmark & GIN    & 9.57  \\
                            & BAShapes & \xmark & \xmark & GTrans & 10.29 \\
                            \cmidrule{2-7}
                            & BAShapes & \checkmark  & \xmark & GCN    & 19.6  \\
                            & BAShapes & \checkmark  & \xmark & GIN    & 19.6  \\
                            & BAShapes & \checkmark  & \xmark & GTrans & 19.6 \\
\bottomrule
\end{tabular}
}
\end{table}

\textbf{Results.} 
 We make the following observations from Table \ref{table:graph-styletransfer-results-bashapes}. First, we observe that using \texttt{GLOBAL-CONTEXT} improves performance on the Sum-Degree and Max-Degree tasks, irrespective of the choice of regressor architecture. However, \texttt{GLOBAL-CONTEXT} is not as helpful for changing the number of triangles. We hypothesize that the LPEs may encode information that makes it difficult to change the number of triangles. Lastly, we note that while the choice of regressor architecture does not significantly impact the performance when \texttt{GLOBAL-CONTEXT} is used, it is influential when they are not used, suggesting that incorporating them by help stabilize training.

\section{Conclusion}
We proposed a novel subgraph-based diffusion model (\ours) to support editing tasks on a single, partially observable network. In particular, \ours uses subgraph sampling and global context to convert single-graph training into learning over a collection of subgraphs and is flexible to choice of diffusion backbone. Notably, this allows us to use expressive GDDM backbones, such as DiGRESS, that were previously prohibited due to quadratic memory requirements. We also formalized three conditional graph editing tasks---namely, expansion, denoising, and style transfer---that can be used for graph refinement with our framework. Lastly, we introduced a new evaluation protocol for our problem setting and performed extensive empirical evaluation to demonstrate the benefits of our proposed framework, as well as the utility of the proposed editing tasks. For future work, we will expand our evaluation to include downstream task performance and our framework to support editing dynamic graphs. 
\bibliography{paper}
\bibliographystyle{iclr2023_conference}
\newpage
\appendix
\onecolumn
\section{Appendix}
\subsection{Acknowledgements}
This work was performed while PT was an intern at Adobe Research. DK is also partially supported by the National Science Foundation under CAREER Grant No.~IIS 1845491, Army Young Investigator Award No.~W9-11NF1810397, and Adobe, Amazon, Facebook, and Google faculty awards. Any opinions, findings, and conclusions or recommendations expressed here are those of the author(s) and do not reflect the views of funding parties. PT thanks Ekdeep Singh Lubana for useful discussions during the course of this project.

\subsection{Denoising Results}
We include our detailed results for denoising here. EDGE uses the degree sequence to help guide the reverse process of large scale graphs. Here, we include a ``cheatcode" that uses the ``target" degree and observed subgraph (instead of the observed degree sequence). In some sense, this provides the complete information needed to recover the unobserved target graph.  
\begin{table}[h!]
\centering
\caption{\textbf{Graph Denoising, BA-Shapes.} 
}
\label{table:graph-denoising-results-bashapes}
\vspace{-2mm}
\footnotesize

\resizebox{\textwidth}{!}{\begin{tabular}{l ccHH ccccc}
\toprule
\textbf{Method} & \textbf{SGDM?} & \textbf{PE?} & \textbf{Alignment?} & \textbf{Subsampling?} & \textbf{Conditioning?} & \textbf{Consensus} $(\uparrow)$ & \textbf{Diversity} $(\uparrow)$ & \textbf{Sparsity} $(\approxeq 1)$ & \textbf{Edge Overlap} $(\approxeq 1)$  
\\
\midrule
EDGE  & \checkmark & \checkmark & \xmark & \checkmark & Obs. Degree + Subst. &  0.256 & 0.903 & 0.874 & 0.8212 \\
EDGE & \checkmark & \checkmark & \xmark & \checkmark & \textcolor{red}{Target} Degree + Subst. & 0.580 & 0.976 & 0.797 & 0.753 \\
\midrule
EDGE & \checkmark & \checkmark & \xmark & \xmark & Obs. Degree + Subst. & 0.341 & 0.928 & 0.762  & 0.721 \\
EDGE & \checkmark & \checkmark & \xmark & \xmark & \textcolor{red}{Target} Degree + Subst. & 0.544 & 0.979 & 0.720 & 0.6838 \\ 
\midrule
EDGE & \xmark & \xmark & \xmark & \xmark & Obs. Degree + Subst. & 0.954 & 0.834 & 0.055 & 0.04273 \\
EDGE & \xmark & \xmark & \xmark & \xmark & \textcolor{red}{Target} Degree + Subst. &  0.950 & 0.852 & 0.063 & 0.04562 \\
\midrule
EDGE & \xmark & \checkmark & \checkmark & \xmark & Obs. Degree + Subst. & 0.996 & 0.733 & 0.019 & 1.519 \\
EDGE & \xmark & \checkmark & \checkmark & \xmark & \textcolor{red}{Target} Degree + Subst. &  0.996 &  0.736 &  0.020 & 1.560 \\
\midrule
GDSS & \checkmark & \xmark & \xmark & \checkmark & Subst. &
0.098 & 0.979 & 0.373 & 0.58165 \\
GDSS & \checkmark & \checkmark & \xmark & \checkmark & Subst. & 0.023 & 0.825 &  0.578 &  0.89488 \\
\midrule
DiGRESS  & \checkmark  &\checkmark & \xmark & \xmark & Subst. & 0.184 &  0.589 & 0.914 & -- \\
DiGRESS   & \checkmark  &\checkmark & \xmark & \checkmark & Subst. & 0.007 & 0.239 & 1.016  & -- \\ 
DiGRESS  & \checkmark  &\checkmark & \xmark & \checkmark & No Subst. & 0.016 & 0.421 & 1.026 & -- \\
\bottomrule
\end{tabular}
}
\end{table}
\begin{table}[h!]
\centering
\caption{\textbf{Graph Denoising, CORA.} 
}
\label{table:graph-denoising-results-cora}
\vspace{-2mm}
\footnotesize

\resizebox{\textwidth}{!}{\begin{tabular}{l ccHH ccccc}
\toprule
\textbf{Method} & \textbf{SGDM?} & \textbf{PE?} & \textbf{Alignment?} & \textbf{Subsampling?} & \textbf{Conditioning?} & \textbf{Consensus} $(\uparrow)$ & \textbf{Diversity} $(\uparrow)$ & \textbf{Sparsity} $(\approxeq 1)$ & \textbf{Edge Overlap} $(\approxeq 1)$  
\\
\midrule
EDGE & \checkmark & \xmark & \checkmark & \xmark & Obs. Degree + Subst. & 0.587 & 1.000 &  0.591 & 0.58509 \\ 
EDGE & \checkmark & \xmark & \checkmark & \xmark & \textcolor{red}{Target} Degree + Subst. & 0.731 & 0.999 &  0.588 & 0.58188 \\
\midrule
EDGE & \checkmark & \checkmark & \xmark & \xmark & Obs. Degree + Subst. & 0.754 &  1.000 &  0.442 & 0.43763 \\
EDGE &  \checkmark & \checkmark & \xmark & \xmark & \textcolor{red}{Target} Degree + Subst. & 0.836 & 1.000 & 0.441 & 0.43677 \\
\midrule
EDGE & \xmark  & \xmark & \xmark & \xmark & Obs. Degree + Subst. & 0.916 & 0.915 & 0.115 &  0.09570 \\
EDGE & \xmark & \xmark & \xmark & \xmark & \textcolor{red}{Target} Degree + Subst. & 0.873 & 0.931 & 0.118 & 0.09387 \\
\midrule
EDGE & \xmark  & \checkmark & \xmark & \xmark & Obs. Degree + Subst. & 0.965 & 0.798 & 0.057 & 0.04864 \\
EGDE  & \xmark & \checkmark & \xmark & \xmark & \textcolor{red}{Target} Degree + Subst. & 0.978 & 0.999 & 0.067 &  0.05109 \\
\midrule
 GDSS & \checkmark & \checkmark & \xmark & \checkmark & Subst. & 0.029 & 0.723 & 0.970 &  0.92559 \\
GDSS & \checkmark & \xmark & \xmark & \checkmark & Subst. &  0.619 &  1.000 & 0.322 & 0.30592 \\ 
\midrule
DiGRESS & \checkmark & \checkmark & \xmark & \checkmark & None & 0.402 & 0.998 & 2.546 & -- \\
DiGRESS & \checkmark & \checkmark & \xmark & \checkmark & Subst. & 0.438 & 0.874 & 0.865 & 0.82835 \\
\bottomrule
\end{tabular}
}
\end{table}
\begin{table}[h!]
\centering
\caption{\textbf{Graph Denoising, PolBlogs.} 
}
\label{table:graph-denoising-results-polblogs}
\vspace{-2mm}
\footnotesize

\resizebox{\textwidth}{!}{\begin{tabular}{l ccHH ccccc}
\toprule
\textbf{Method} & \textbf{SGDM?} & \textbf{PE?} & \textbf{Alignment?} & \textbf{Subsampling?} & \textbf{Conditioning?} & \textbf{Consensus} $(\uparrow)$ & \textbf{Diversity} $(\uparrow)$ & \textbf{Sparsity} $(\approxeq 1)$ & \textbf{Edge Overlap} $(\approxeq 1)$  
\\
\midrule
EDGE  & \checkmark & \xmark & \xmark & \xmark & Obs. Degree + Subst. &  0.713 &  1.000 &  0.340 &  0.31515  \\
EDGE & \checkmark & \xmark & \xmark & \xmark & \textcolor{red}{Target} Degree + Subst. & 0.605 & 1.000 & 0.437 &  0.4044 \\  
\midrule
EDGE & \checkmark & \checkmark & \xmark & \xmark & Obs. Degree + Subst. &  0.375 & 1.000 &  0.640 & 0.59266 \\
EDGE & \checkmark & \checkmark  & \xmark & \xmark & \textcolor{red}{Target} Degree + Subst. & 0.469 &  1.000 &  0.686 & 0.6356 \\
\midrule
EDGE & \xmark & \checkmark & \xmark & \xmark & Obs. Degree + Subst. & 0.993 &  0.598 & 0.023 & 0.02181 \\
EDGE & \xmark & \checkmark & \xmark & \xmark & \textcolor{red}{Target} Degree + Subst. &  0.991 &  0.539 & 0.018 & 0.0170   \\
\midrule
EDGE & \xmark & \xmark & \xmark & \xmark & Obs. Degree + Subst. & 0.992 & 0.702 & 0.0411 & 0.0412 \\
EDGE & \xmark & \xmark & \xmark & \xmark & \textcolor{red}{Target} Degree + Subst. &  0.992 &  0.717 & 0.042 & 0.03972 \\
\midrule
GDSS & \checkmark & \xmark & \xmark & \checkmark & Subst. &  0.683 & 1.000 &  0.353 & 0.31975 \\
GDSS & \checkmark & \checkmark & \xmark & \checkmark & Subst. & 0.010 & 0.605 & 1.078 & 0.97685 \\
\midrule
DiGRESS   & \checkmark  &\checkmark & \xmark & \checkmark & Subst. &  0.8239 &  0.99603 & 0.60938  & -- \\ 
DiGRESS & \checkmark & \checkmark & \xmark  & \checkmark & No Subst. & 0.886 & 1.000 &  2.125 & 0.42513 \\
\bottomrule
\end{tabular}
}
\end{table}
\newpage

\subsection{Expansion Results}
We include our detailed results for expansion here. EDGE uses the degree sequence to help guide the reverse process of large scale graphs. Here, we include a ``cheatcode" that uses the ``target" degree and observed subgraph (instead of the observed degree sequence). In some sense, this provides the complete information needed to recover the unobserved target graph.  
\begin{table}[h!]
\centering
\caption{\textbf{Graph Expansion, BA-Shapes.} 
}
\label{table:graph-expansion-results-ba}
\vspace{-2mm}
\footnotesize
\resizebox{\textwidth}{!}{\begin{tabular}{l ccH ccccc}
\toprule
 \textbf{Method} & \textbf{SGDM?} & \textbf{PE?} & \textbf{Alignment?} & \textbf{Conditioning?} & \textbf{Consensus} $(\uparrow)$ & \textbf{Diversity} $(\uparrow)$ & \textbf{Sparsity} $(\approxeq 1)$ & \textbf{Edge Overlap} $(\approxeq 1)$  
\\
\midrule
EDGE & \xmark & \checkmark & \xmark & Obs. Degree + Subst & 0.000 & 0.100 & 0.930 & 1.000 \\
EDGE & \xmark & \checkmark & \xmark & \textcolor{red}{Target} Degree + Subst. & 0.044  & 0.618 & 0.958  &1.000 \\
\midrule
EDGE & \xmark & \xmark & \xmark & Obs. degree + Subst. & 0.000 & 0.100 & 0.930 & 1.000 \\ 
EDGE & \xmark & \xmark & \xmark & \textcolor{red}{Target} Degree + Subst. & 0.030 & 0.598 & 0.960 & 1.000 \\

\midrule
EDGE & \checkmark & \xmark & \xmark & Obs. Degree + Subst. & 0.055 & 0.100 & \underline{0.930} & 1.000 \\ 
EDGE & \checkmark & \xmark & \xmark & \textcolor{red}{Target} Degree + Subst. & 0.589 & 0.511 & 1.002 & 1.000 \\ 
\midrule
EDGE & \checkmark & \checkmark &  \xmark & Obs. Degree + Subst. & 0.055 &  0.100 & \underline{0.930} & 1.000 \\ 
EDGE & \checkmark & \checkmark & \xmark & \textcolor{red}{Target} Degree + Subst. & 0.594 &  0.511 & 1.002 & 1.000 \\ 
\midrule
GDSS & \checkmark & \xmark & \xmark & Subst. & 0.296 & 0.983 & 1.707 & 1.000 \\
GDSS & \checkmark & \checkmark & \xmark &  Subst. & 0.869 & 0.565 & 8.798 & 1.000\\
\midrule
DiGRESS & \checkmark & \checkmark & \xmark & Subst. & \underline{0.309} & 0.616 & \textbf{1.039} & 1.000 \\
DiGRESS & \checkmark & \checkmark & \xmark & None & \textbf{0.550} & \textbf{1.000} & 5.738 & 1.000  \\
\bottomrule
\end{tabular}
}
\end{table}

\begin{table}[h!]
\centering
\caption{\textbf{Graph Expansion, CORA.}
}
\label{table:graph-expansion-results-cora}
\vspace{-2mm}
\footnotesize

\resizebox{\textwidth}{!}{\begin{tabular}{l ccH ccccc}
\toprule
 \textbf{Method} & \textbf{SGDM?} & \textbf{PE?} & \textbf{Alignment?} & \textbf{Conditioning?} & \textbf{Consensus} $(\uparrow)$ & \textbf{Diversity} $(\uparrow)$ & \textbf{Sparsity} $(\approxeq 1)$ & \textbf{Edge Overlap} $(\approxeq 1)$  
\\
\midrule
EDGE & \xmark & \xmark & \xmark & Obs. Degree + Subst. & 0.000 & 0.100 & 0.922 & 1.00 \\  
EDGE & \xmark & \xmark &  \xmark & \textcolor{red}{Target} Degree + Subst. & 0.020 & 0.520 & 0.944 & 1.00 \\  
\midrule
EDGE & \xmark & \checkmark  & \xmark & Obs. Degree + Subst. & 0.000 & 0.100 & 0.922 & 1.000\\  
EDGE & \xmark & \checkmark  &  \xmark & \textcolor{red}{Target} Degree + Subst. & 0.020 &  0.493 & 0.940 &  1.000 \\  
\midrule
EDGE & \checkmark & \xmark & \xmark & Obs. Degree + Subst. & 0.000 & 0.100 & 0.925 & 1.00  \\  
EDGE & \checkmark & \xmark & \xmark & \textcolor{red}{Target} Degree + Subst. & 0.550 & 0.510 & 1.001 & 1.00 \\ 
\midrule
EDGE & \checkmark & \checkmark & \xmark & Obs. Degree + Subst. & 0.000 & 0.100 & 0.918 & 1.000 \\ 
EDGE & \checkmark & \checkmark & \xmark & \textcolor{red}{Target} Degree + Subst. & 0.542 & 0.535 & 1.000 & 1.000 \\ 
\midrule
GDSS & \checkmark & \xmark & \xmark & Subst. & 0.246 & 1.000 & 5.098 & 1.000 \\
GDSS & \checkmark & \checkmark & \xmark & Subst. & 0.223 & 1.000 & 10.730 & 1.000 \\
\midrule
DiGRESS & \checkmark & \checkmark & \xmark & Subst. & 0.117 & 0.933 &  1.831 & 1.000 \\
DiGRESS & \checkmark & \checkmark & \xmark  & None & 0.142 & 0.968 &  1.755 & 1.000\\  
\bottomrule
\end{tabular}
}
\end{table}
\begin{table}[h]
\centering
\caption{\textbf{Graph Expansion, PolBlogs.}}
\label{table:graph-expansion-results-polblogs}
\vspace{-2mm}
\footnotesize
\resizebox{\textwidth}{!}{\begin{tabular}{l ccH ccccc}
\toprule
\textbf{Method} & \textbf{SGDM?} & \textbf{PE?} & \textbf{Alignment?} & \textbf{Conditioning?} & \textbf{Consensus} $(\uparrow)$ & \textbf{Diversity} $(\uparrow)$ & \textbf{Sparsity} $(\approxeq 1)$ & \textbf{Edge Overlap} $(\approxeq 1)$  
\\
\midrule
EDGE	&	\xmark	&	\xmark	&	\xmark	& Obs. Degree + Subst.				&	0	&	0.100 & 0.941 & 1.000\\	
EDGE	&	\xmark	&	\xmark	&	\xmark	& \textcolor{red}{Target}	Degree	+	Subst.	&	0.046	&	0.909 & 0.998 & 1.000\\	
\midrule
EDGE	&	\xmark	&	\checkmark	&	\xmark	& Obs. Degree + Subst.				&	0	&	0.100 & 0.938 & 1.000 \\	
EDGE	&	\xmark	&	\checkmark	&	\xmark	&\textcolor{red}{Target}	Degree	+	Subst.	&	0.046	&	0.888 &  0.994 &  1.000 \\	
\midrule
EDGE & \checkmark & \xmark & \xmark & Obs. Degree + Subst. & 0.000 &  0.100 & 0.939 & 1.000 \\  
EDGE & \checkmark & \xmark & \xmark & \textcolor{red}{Target} Degree + Subst.  & 0.681 &  0.363 &  1.002 & 1.000\\  
\midrule
EDGE & \checkmark & \checkmark & \xmark & Obs. Degree + Subst. & 0.000 & 0.100 & 0.939 & 1.000\\  
EDGE & \checkmark & \checkmark & \xmark & \textcolor{red}{Target} Degree + Subst. & 0.652 & 0.392 & 1.002 & 1.000 \\  
\midrule
GDSS & \checkmark & \xmark & \xmark &  Subst. & 0.217 & 1.000 & 5.444 & 1.000\\  
GDSS & \checkmark & \checkmark & \xmark &  Subst. & 0.305 & 1.000 & 14.174 & 1.000 \\  
\midrule
DiGRESS & \checkmark & \checkmark & \xmark &  Subst.  & 0.034 & 0.991 & 1.153 & 1.000 \\  
DiGRESS & \checkmark & \checkmark & \xmark & None & 0.031 & 0.994 & 1.134 & 1.000 \\  
\bottomrule
\end{tabular}
}
\end{table}

\newpage
\subsection{Style Transfer Results}
\begin{table}[h]
\centering
\caption{\textbf{Graph Style Transfer}}
\label{table:graph-styletransfer-results-polblogs}
\vspace{-2mm}
\footnotesize
\resizebox{0.5\textwidth}{!}{\begin{tabular}{lHcHccH}
\toprule
\textbf{Task} & \textbf{Task} & \textbf{PE?} & \textbf{Conditioning?} & \textbf{Regressor Type} & \textbf{MAE}$(\downarrow)$ & \textbf{Lamda} \\ \midrule
\multirow{6}{*}{Sum-Degree} & BAShapes & \xmark & \xmark & GCN    & 6.46  \\
                            & BAShapes & \xmark & \xmark & GIN    & 6.81  \\
                            & BAShapes & \xmark & \xmark & GTrans & 6.48  \\
                            \cmidrule{2-7}
                            & BAShapes & \checkmark  & \xmark & GCN    & 3.78  \\
                            & BAShapes & \checkmark  & \xmark & GIN    & 3.62  \\
                            & BAShapes & \checkmark  & \xmark & GTrans & 3.45  \\
\midrule
\multirow{6}{*}{Max-Degree} & BAShapes & \xmark & \xmark & GCN    & 24.73 \\
                            & BAShapes & \xmark & \xmark & GIN    & 45.09 \\
                            & BAShapes & \xmark & \xmark & GTrans & 24.63 \\
                            \cmidrule{2-7}
                            & BAShapes & \checkmark  & \xmark & GCN    & 9.3   \\
                            & BAShapes & \checkmark  & \xmark & GIN    & 11.04 \\
                            & BAShapes & \checkmark  & \xmark & GTrans & 11.44 \\
\midrule
\multirow{6}{*}{Num. Triangles}  & BAShapes & \xmark & \xmark & GCN    & 9.18  \\
                            & BAShapes & \xmark & \xmark & GIN    & 9.57  \\
                            & BAShapes & \xmark & \xmark & GTrans & 10.29 \\
                            \cmidrule{2-7}
                            & BAShapes & \checkmark  & \xmark & GCN    & 19.6  \\
                            & BAShapes & \checkmark  & \xmark & GIN    & 19.6  \\
                            & BAShapes & \checkmark  & \xmark & GTrans & 19.6 \\
\bottomrule
\end{tabular}
}
\end{table}

\newpage
\subsection{Experimental Figures}



\begin{figure}[h]
    \centering
    \includegraphics[width=1\linewidth]{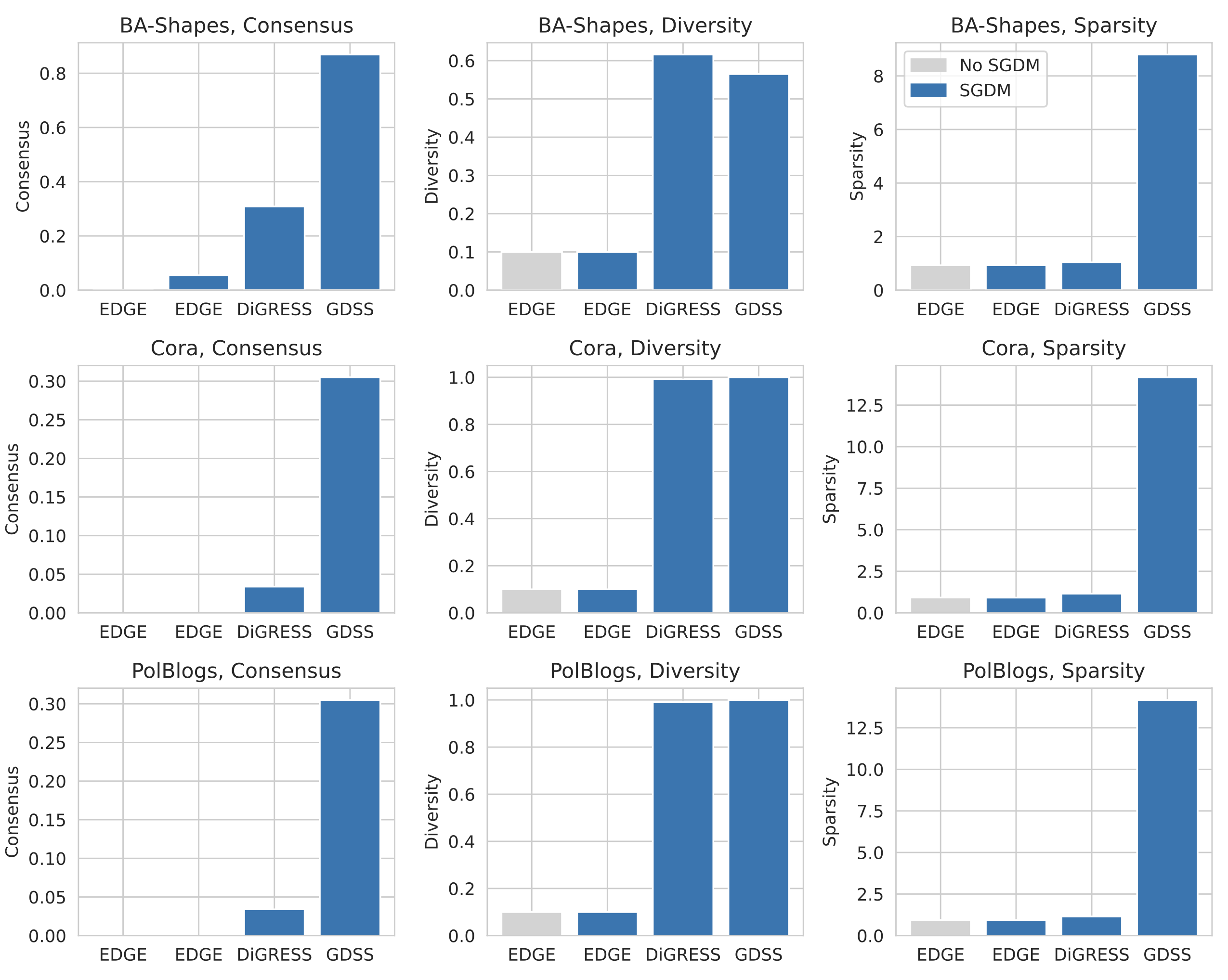}
    \caption{\textbf{Effect of Diffusion Backbone on Denoising}}
    \label{fig:denoising_choice_of_diffusion_ex}
\end{figure}

\begin{figure}[h]
    \centering
    \includegraphics[width=1\linewidth]{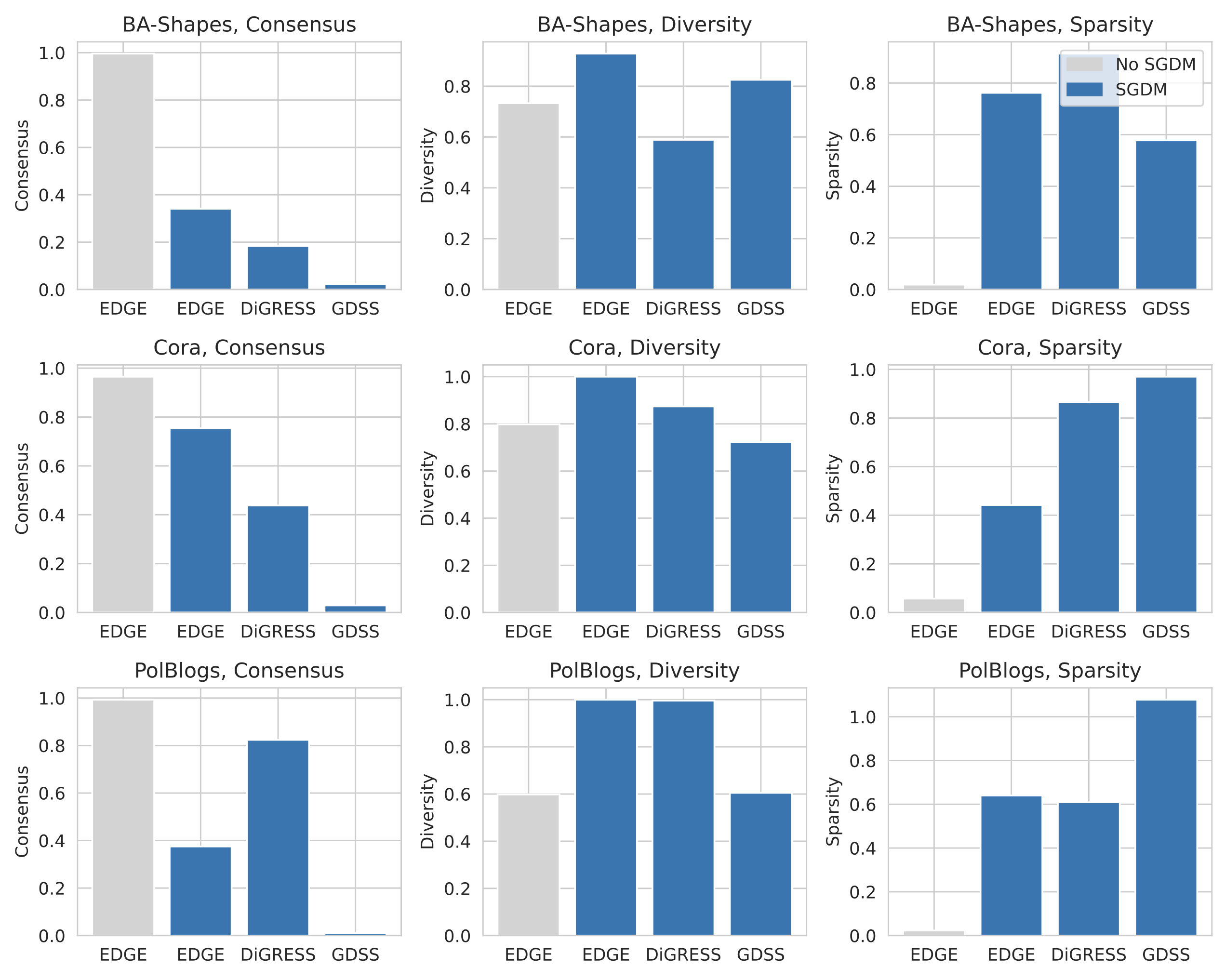}
    \caption{\textbf{Effect of Diffusion Backbone on Expansion.} Not that while it appears that No SGDM-EDGE has better consensus, note that it also has significantly lesser sparsity. }
    \label{fig:denoising_choice_of_expansion_ex}
\end{figure}

\newpage
\subsection{Generating Large Graphs with Subgraph-based Diffusion}
Here, we demonstrate that our proposed subgraph-based diffusion framework performs comparably with other generative graph models. We use a simple ``stitching" based algorithm: (1) we generate an ego-graph around a particular node, and (2) stitches the subgraphs together. Positional encodings are used in lieu of node-ids as the positional encodings of additional nodes can be approximated using perturbation theory. 

\begin{algorithm}[h]
\small
\caption{\small Large Graph Generation with \ours}
\label{algo:sgdm_stitching}
\KwIn{$p_\theta$:trained reverse process, $\texttt{Hist-Global}$: distribution of observed global context, $\texttt{Hist-GraphSize}$: distribution of observed subgraph sizes}
\KwOut{$\gengraph = (\mA_G, \mX_G, \mC_G,\mQ_G)$: generated graph}
$\bm{\mathcal{S}} = \{\}$\\
\While{$\exists C_{i} \in \mC$ not SAMPLED}{
$n \gets \texttt{SAMPLE}(\texttt{Hist-GraphSize})$ \\
$\mC_{\mathcal{S}_T} \gets \texttt{SAMPLE}(\texttt{Hist-Global},n)$ \\
$\mathcal{S}^{t=T} \gets q_n(\cdot)$ \\
\For{t = T \KwTo 0}{
    $\mQ^t \gets \texttt{Local-Context}(\mathcal{S}^t)$ \\
    $\mathcal{S}^t \gets \mQ^t $ \\ 
    $\mathcal{S}^{t-1} \gets p_\theta(\mathcal{S}^t,\mC)$ \\
    $ (\mA^{t-1}, \mX^{t-1},\mC,\mQ^t) \gets \mathcal{S}^{t-1}$ \\
}
$\bm{\mathcal{S}}\gets \mathcal{S}^{t=0}$ \\
}
\Comment{Use Global-Context to Merge Subgraphs.} 
$\mathcal{S}_G = \texttt{COALESCE}(\bm{\mathcal{S}})$ \\
return $\mathcal{S}_G$
\end{algorithm}

\begin{figure}[h]
\vspace{-0.4cm}
    \centering
    \includegraphics[width=.8\textwidth]{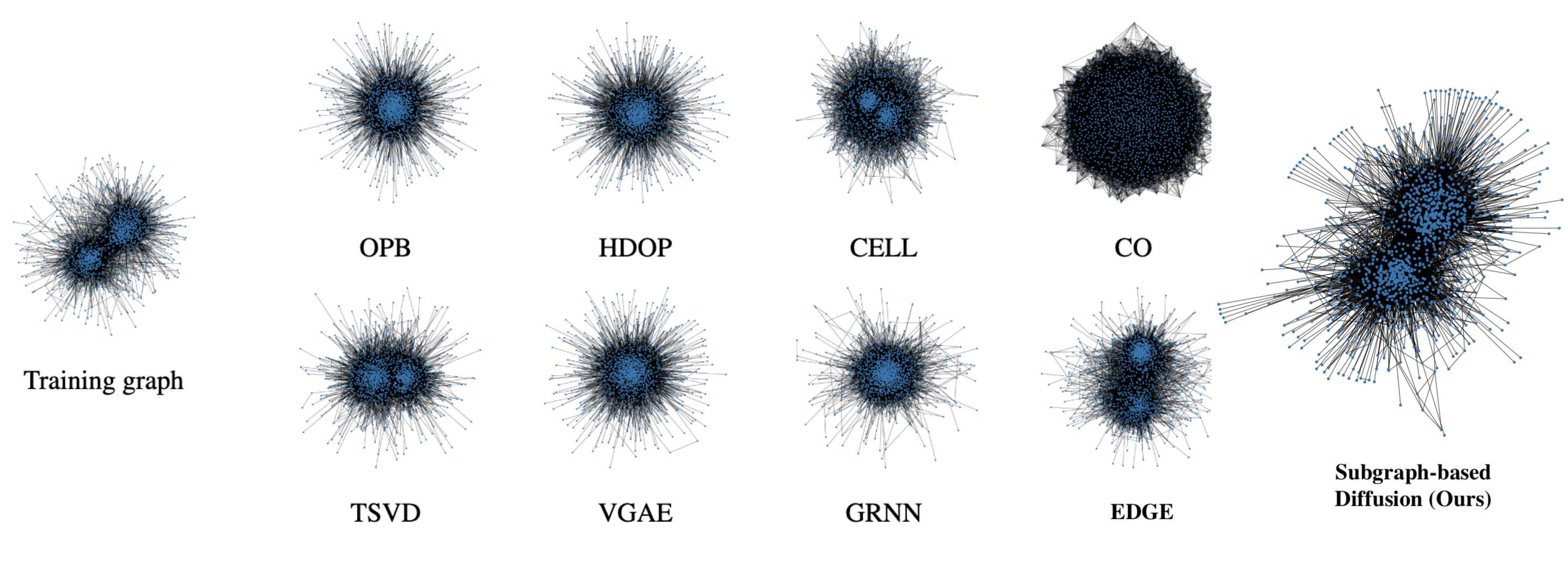}
    \vspace{-0.4cm}
    \caption{\textbf{Example of Stitching.} Compared against several other large-scale generative models, including auto-encoder, recurrent and diffusion based models, we see that our proposed subgraph-based diffusion framework is able to generate graphs qualitatively similar graphs to the training sample. Visualization of other methods is reproduced from \cite{Chen23_EDGE}.}
    \label{fig:enter-label}
\end{figure}
\begin{table*}[h]
    \centering
    \small
\caption{\textbf{Large Graph Generation using Subgraph Diffusion.} On the PolBlogs dataset, we compare the performance of our proposed subgraph diffusion framework against several large graph generation methods. The performance of other generative models is reproduced from \cite{Chen23_EDGE}. Better performance is obtained models which produce graphs whose statistics are closer to the training graph, without having memorized the training graph (Edge Overlap $<< 100$).} 
\vspace{-0.2cm}

\resizebox{\textwidth}{!}{\begin{tabular}{lcccccc}\toprule
 & \textbf{Edge Overlap} & \textbf{Power Law Exp.} & \textbf{Num. Triangle} & \textbf{Clustering Coeff.} & \textbf{Characteristic Path Length} & \textbf{Assortativity Coeff.} 

\\\midrule
Training Graph & 
100 & 1.414 & 1 & 0.226 & 2.738 & -0.221 

\\
OPB \citep{Chanpuriya21_CCOP} & 
24.5 & 1.395 & 0.667 & 0.150 & 2.524 & -0.143 

\\
HDOP \citep{Chanpuriya21_CCOP} & 
16.4 & 1.393 & 0.687 & 0.153 & 2.522  & -0.131 

\\ 
CELL \citep{Rendsburg20_CELL} & 
26.8 & 1.385 & 0.810 & 0.211 & 2.534 & -0.230 

\\
CO \citep{Chanpuriya21_CCOP} & 
20.1 & 1.975 & 0.045 & 0.028 & 2.502 & 0.068 

\\ 
TSVD \citep{Seshadhri20_TSVD} & 
32.0 & 1.373 & 0.872 & 0.205 & 2.532  & -0.216 
\\
VGAE \citep{Kip16_VGAE} & 
3.6 & 1.723 & 0.05 & 0.001 & 2.531 & -0.086 
\\
GRNN \citep{You18_GraphRNN} & 
9.6 & 1.333 & 0.354 & 0.095 & 2.566 & 0.096 
\\
EDGE \citep{Chen23_EDGE} & 
16.5 & 1.398 & 0.977 & 0.217 & 2.647 & -0.214 
\\
\textbf{SGDM(Ours)} & 
9.44 & 1.384 & 1.301	& 0.308 & 2.612 & -0.264
\\
\bottomrule
\end{tabular}
}
\label{tab:main-large-network}
\end{table*}

\newpage
\subsection{Expanded Related Work}\label{app:expanded_related_work}
Here, we expand on the related work introduced in Sec. \ref{sec:related-work}. In brief, diffusion-based generative models consist of two parts: (i) a \textit{forward} process, in which data, $\vx_0 \sim q(\vx)$, is gradually transformed to a reference distribution through the addition of some noise over time $\epsilon_t$, and (ii) \textit{reverse} process, which takes samples from the reference distribution and attempts to denoise the samples over time to synthesize samples from the original data distribution.


Early graph diffusion models closely followed the formulation of diffusion models for image synthesis, and ignored the discreteness of graph data when designing these processes. For example, \citet{Niu20_EDPGNN} use a multivariate standard Gaussian as the reference distribution and add Gaussian noise to the adjacency matrix when the performing the forward step. After the final reverse step is completed, the continuous graph is discretized. While their discrete-time score-based formulation leads to permutation invariant generation (e.g., different node permutations of a graph have the same probability), intermediate representations are continuous, reducing their interpretability and requiring 1000+ steps to generate good samples. \citet{Jo22_GDSS} propose a continuous-time system of stochastic differential equations to jointly capture the dependency of nodes and graph topology. However, since the forward process still adds Gaussian noise to the adjacency matrix, it also suffers from cumbersome sampling. 

A Gaussian reference distribution and noising process was well-suited for image synthesis, as it not only enabled tractable formulations of the forward/reverse processes, but also supported natural image inductive biases. Namely, that adding Gaussian noise in the forward process will remove high frequency components (style) before low frequency components (content), allowing models to learn to ``paint" pictures. Because graphs are discrete, structured data, Gaussian noise/priors are not as applicable since they destroy the sparsity and alter topological properties, leading to subpar performance. Therefore, more recent graph diffusion models have focused on the discrete diffusion processes \citep{Austin21_StructuredDenoiseing,Hoogeboom21_ArgmaxFlows}.

Instead of using multivariate Gaussian's as the reference distribution and Gaussian noise during the forward process, \citet{digress_vignac22} uses a uniform distribution over different types of edge and nodes types as the reference distribution and performs edge/node edits to simulate changing the particular edge/node category. Since the reference distribution is over the categorical edge/node, they cast the training objective as correctly classifying the edge/node type as each time step, which is much simpler than learning each node/edge. Moreover, they find that defining the transition probabilities in terms of the marginal distribution over the edge/node anchors observed in the training distribution (instead of uniform) can substantially decrease the number of denoising steps and improve the overall generation quality. Lastly, to circumvent limitations in GNN expressiveness, DIGRESS leverages the discrete intermediate steps to compute structural features at each time step. However, we note that DiGRESS is expensive to train on large graphs and expects only categorical edges/nodes, which can be limiting for real-datasets. 

In order to improve the scalability of graph diffusion models, while retaining high-fidelity, \citet{Chen23_EDGE} (EDGE) proposes an alternative discrete denoise process, which uses an empty graph as the reference distribution and defines the forward process as sequentially removing edges to an empty state. In order to effectively scale to large graphs, EDGE limits the number of ``active" nodes by only conditioning the propagation step on whether a nodes degree has changed from $t$ to $t-1$. Furthermore, since the degree distribution is closely related to many graph properties, EGDE conditions on the degree distribution (since they use a discrete process), a tangential benefit of their discrete process. \citet{Jo23_DestinationDriven} (DRuM) learns distribution over possible trajectories using Schrodinger's Bridges, instead of the a single point, in order to ensure that intermediate steps are well-aligned with the true data-distribution. Lastly, we note that our work is contemporary to \citet{Limios23_SaGESS}, which also uses sampling and reconstruction to improve the scalability of diffusion models (in particular DiGress). However, there are several differences: namely, SGDM is a framework that is flexible to the choice of underlying diffusion model, uses both global/local contexts to improve the performance, and does not require discrete node feature ids to perform the reconstruction step. Furthermore, in addition to SGDM, we also formalize several network editing tasks and demonstrate that SGDM can be used to effectively perform these tasks.  

\newpage
\subsection{Experimental Details}\label{app:experimental_details}
We trained all models using Tesla T4s (16GB GPU Memory, 124GB RAM). To ensure fair comparison across methods and prevent overfitting to a corrupted graph, all models are trained for at most 24 hours or 5000 epochs, which ever came first. All metrics are reported over 10 generated samples. When computing edge overlap, we first sort the graph by the degree sequence and then compare. 

We use the official released code by DiGRESS (\url{https://github.com/cvignac/DiGress}), EDGE (\url{https://github.com/tufts-ml/graph-generation-EDGE}) and GDSS (\url{https://github.com/harryjo97/GDSS}). Our evaluation set-up, metrics and editing tasks were implemented using PyTorch Geometric~\citep{Fey19_PyG} and PyTorch. Our code can be accessed at (\url{https://anonymous.4open.science/r/DiGress-PT-781D/},\url{https://anonymous.4open.science/r/EDGE-PT-528C/} , \url{})

For the \texttt{SAMPLING} function, we use 2-hop ego networks and, for \texttt{SUBSAMPLING}, perform uniform randomly subsampling to obtain subgraphs of maximum graph-size 50 for training. For \texttt{GLOBAL-CONTEXT}, we use the top-two non-zero eigenvectors of \targetgraph~'s Laplacian. Hyper-parameters and architectures suggested by each method's authors are used. However, to incorporate global context, we use the following strategy, inspired by how vision diffusion models incorporate positional coordinates~\citep{Luhman22_ImprovingDiffusion,wang23_PatchDiffusion}: (i) we project the global context vector to the same size of the node's hidden representation vector (ii) at each message passing layer, we add the global context projection to each node representation.

As discussed in Sec. \ref{sec:related-work} and Sec. \ref{app:expanded_related_work}, EDGE uses the degree sequence to help guide large models in the reverse process. We include additional refinement results that show the performance of EDGE when using the observed subgraph and the \textit{target} degree sequence. This can be seen as giving the reverse process a ``cheat-code", since all the information to recover the unobserved, clean subgraph has been provided. We note that using EDGE directly with the observed subgraph's degree distribution (without global-context) requires that we perform graph-alignment with respect to the observed subgraph. Using global context with the observed degree helps avoid this alignment issue for feature-less graphs as ids can be matched using the global context. While we directly filled in (removed) edges when working with DiGRESS, GDSS uses a real-valued adjacency matrix. Therefore, when performing expansion, we would fill in the maximum value at a given time $t$ for the edges that we wished to condition upon. Lastly, we note that while our expansion task definition appears similar to link prediction, this task and the discuss in-painting approach also provides support for filling in missing features. Since EDGE does not support features at this time, we only present results for subgraph denoising. 


\begin{table}
\footnotesize
\centering
\caption{\label{table:subgraph_stats} \textbf{\texttt{SAMPLED} Dataset Statistics.} After performing 2-hop ego net sampling, we obtain the following dataset statistics. Note, that while the average and median number of edges is large for PolBlogs, we perform \texttt{SUBSAMPLING} to control this size. Furthermore, note that since average graph size is now more mangeable, we can also perform batching to reduce training time.}
\begin{tabular}{l cccccc}
\toprule
\textbf{Dataset} & \textbf{Num. Graphs} & \textbf{Avg. Nodes} &  \textbf{Med. Nodes} & \textbf{Avg. Edges} & \textbf{Med. Edges}
\\
\midrule
BA-Shapes & 700 &  159 & 91 & 1,300 & 502\\
PolBlogs  & 1,222 &  486 & 519 & 15,768 & 17,129 \\
CORA & 2,485 &  139 & 92 & 524 & 296  \\
\bottomrule
\end{tabular}%
\end{table}

\newpage
\subsection{Alternative Instantiations}\label{app:sgdm_details}
We used $k$-hop ego networks for the \texttt{SAMPLING} functions as we desired a node-centric sampling scheme and induce subgraphs from a randomly selected subset of larger graphs to perform \texttt{SUBSAMPLING}. We were motivated from the perspective that particular nodes may be unreliable and lead to corruptions in the observed graph. For example, we can imagine malicious users in a social network. Given that the majority of nodes (and their surrounding neighborhoods) are not corrupted and nodes (users) are expected to have some similar behaviors, learning a distribution over the induced ego-centric subgraphs will allow information transfer between the majority, uncorrupted nodes and the corrupted ones.  

While we use random subsampling and ego-nets in our experiments, alternatives are also possible ~\citep{Leskovec06_LargeGraph}. 
For example, instead of effectively doing Breadth First Search, we can also do random walk sampling, forest fire sampling, or potentially node2vec style sampling starting from each node in the observed graph so that we can balance the amount of global network structure or local network structure captured in each subgraph. While the above provide different strategies for generating node-centric graphs, we can also define a \texttt{SAMPLING} function that is not node-centric but still satisfies the criteria introduced in Sec. \ref{sec:approach}. For example, we can use METIS to obtain a non-overlapping graph partition and extend each partition by 1-hop to make sure to share information across partitions. The choice of \texttt{SAMPLING} function helps control how many times a single node is seen over the collection of all subgraphs. 

We note that the choice of \texttt{SUBSAMPLING} function is also flexible in \ours. While we used uniform node sampling for its ease of use, it is also possible to perform \texttt{SUBSAMPLING} with other strategies. For example, we can use a distribution biased by node degree instead of a uniform distribution. Lastly, we note that there are several choices for \texttt{GLOBAL-CONTEXT} as well. For example, instead of Laplacian Positional vectors, we can use position aware embeddings~\citep{You19_positionawareGNN}, which computes a distance between a node and a set of anchor nodes.

\newpage
\subsection{Additional Experimental Results}\label{app:extra_experiments}

\begin{figure}[h]
    \centering
    \includegraphics[width=0.3\textwidth]{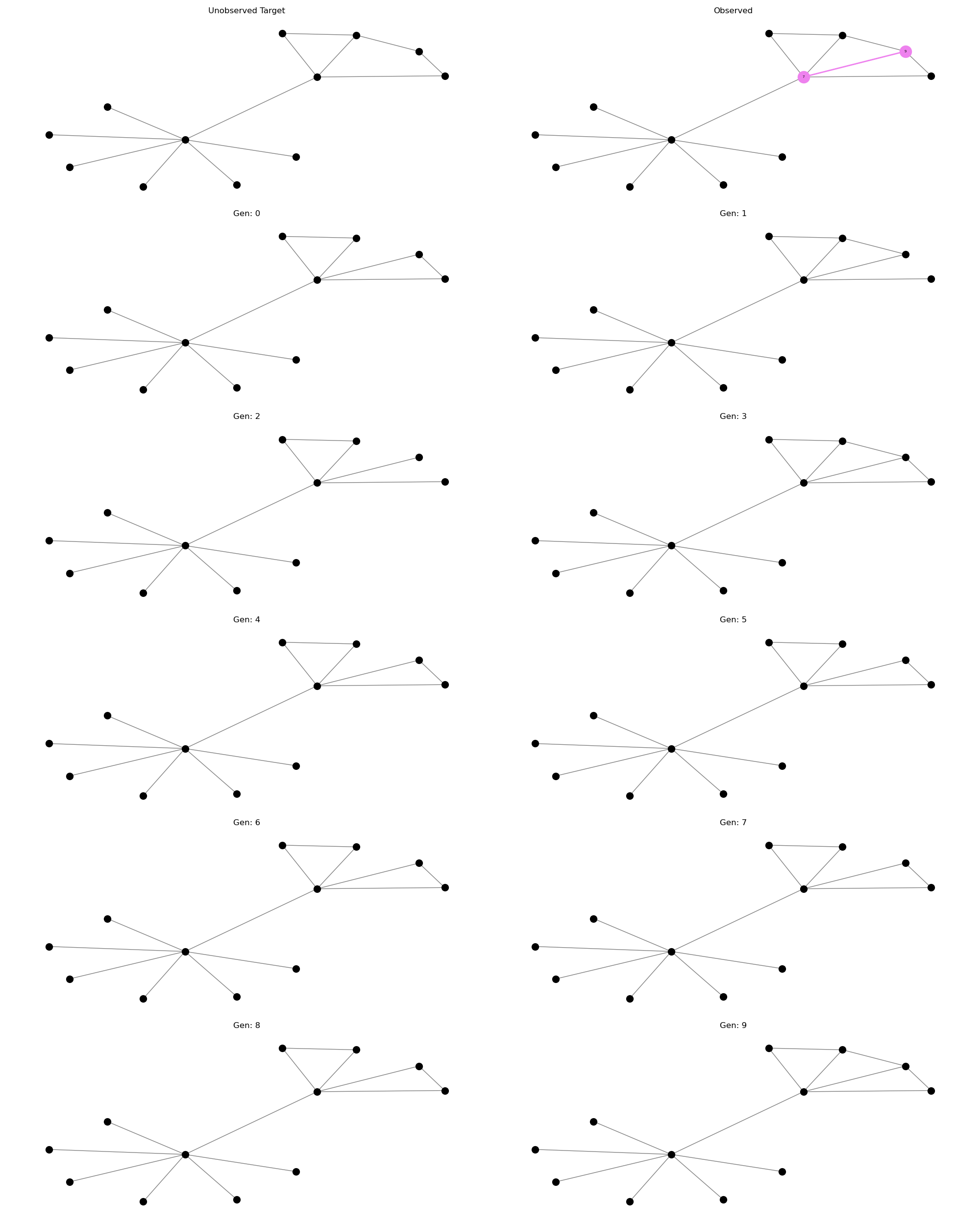}
    \includegraphics[width=0.3\textwidth]{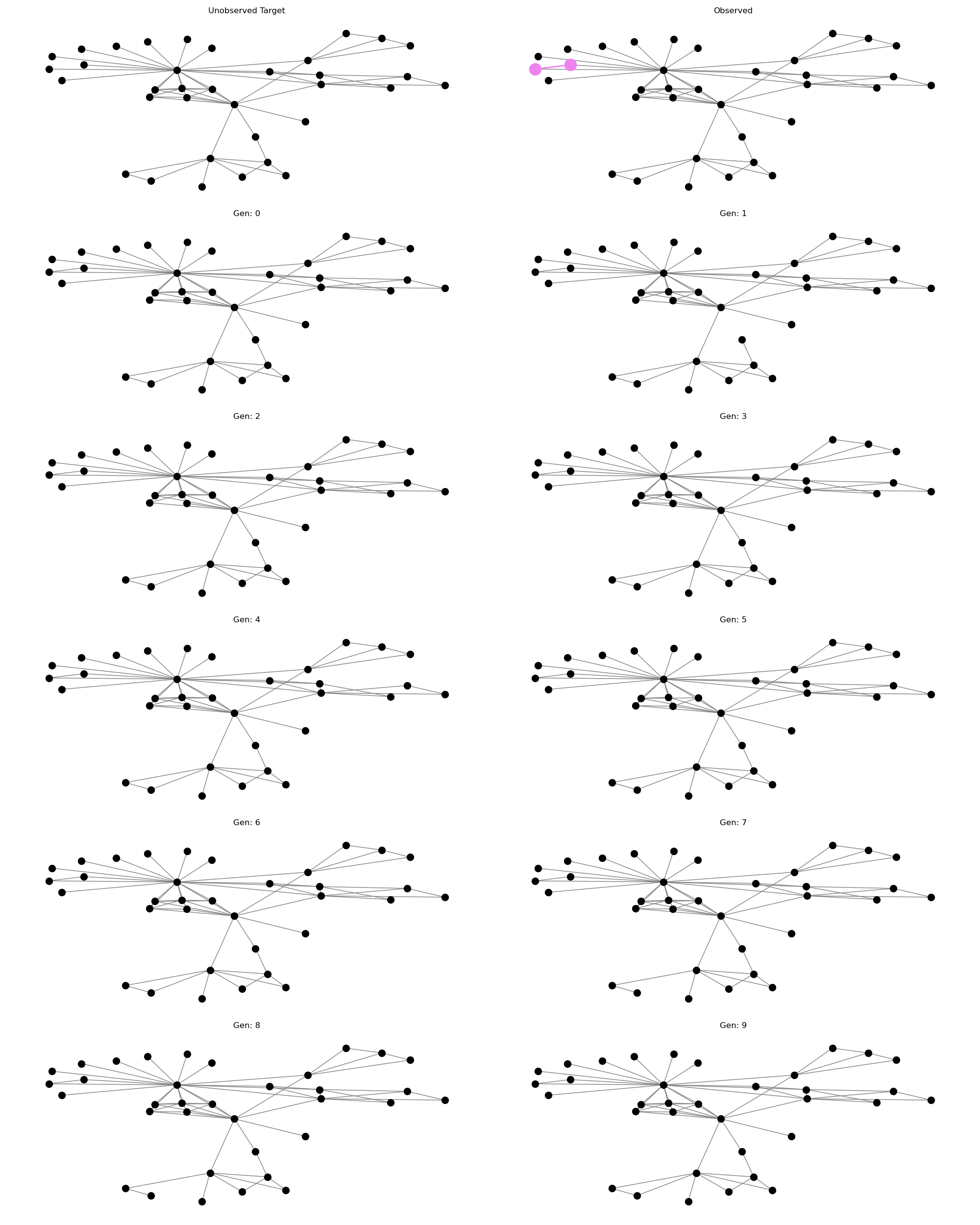}
    \includegraphics[width=0.3\textwidth]{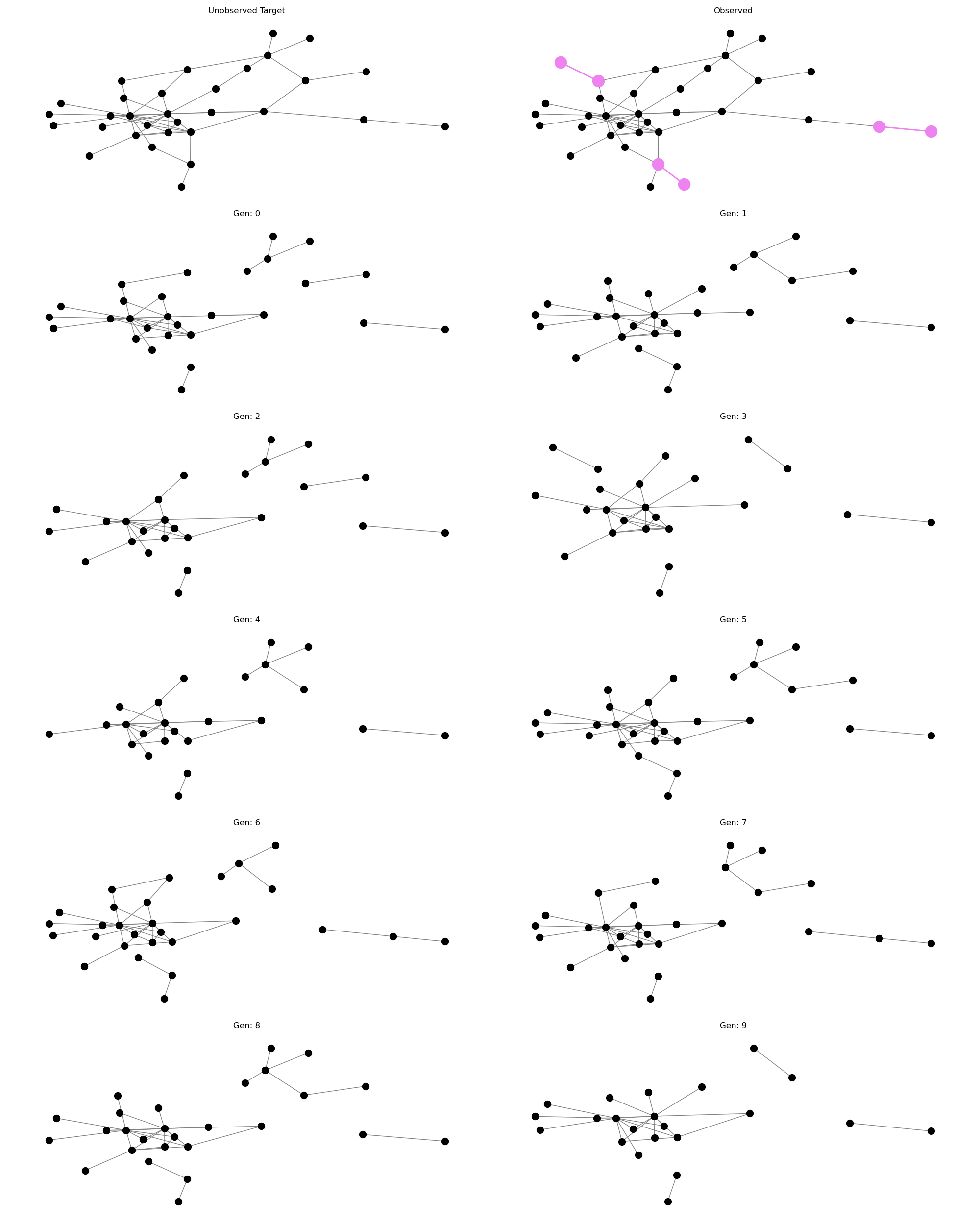}
    \label{fig:digress_bashapes_denoising_ex}
    \caption{\textbf{Digress, (BAShapes, Cora, PolBlogs) Denoising.}}
\end{figure}

\begin{figure}[h]
    \centering
    \includegraphics[width=0.3\textwidth]{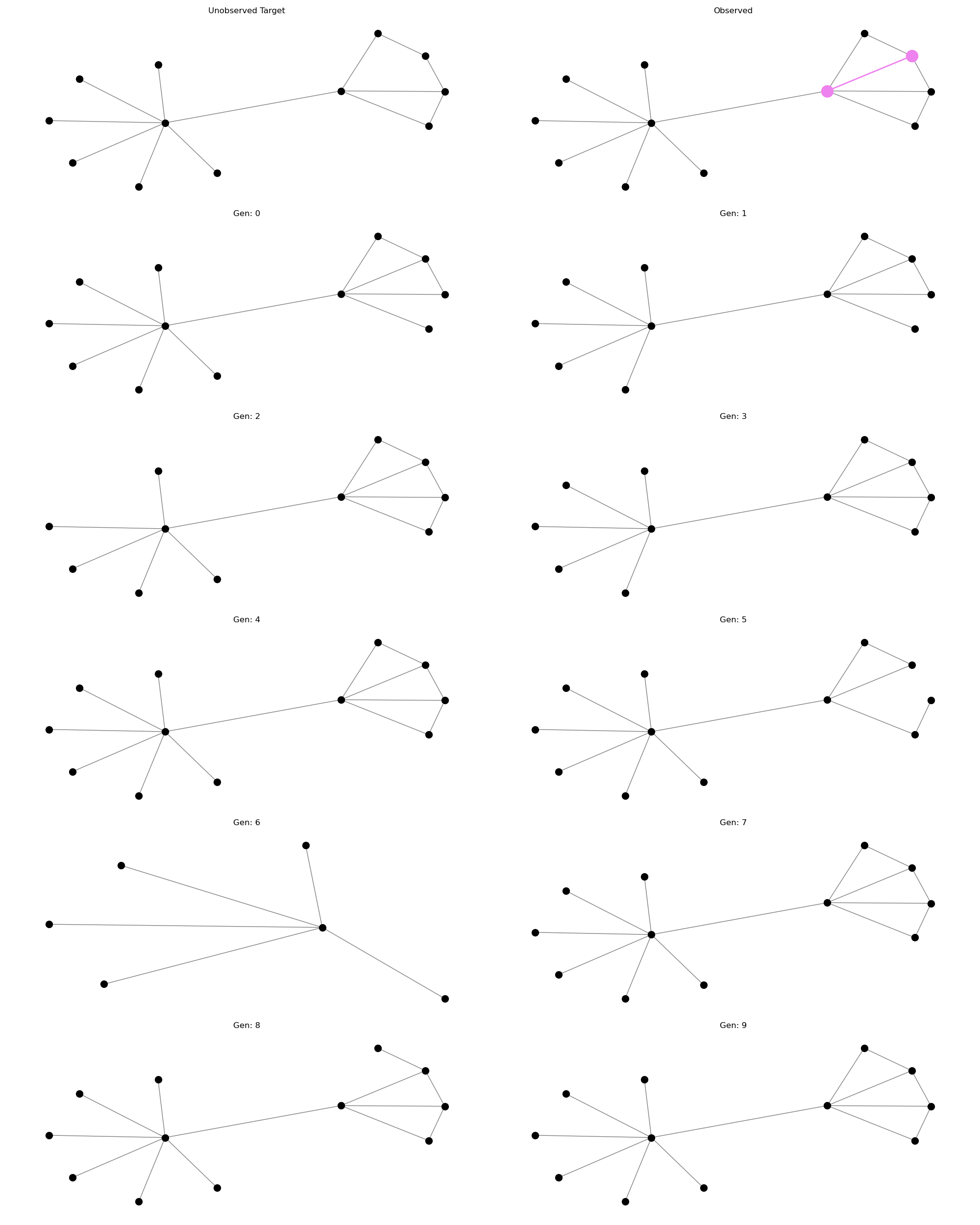}
    \includegraphics[width=0.3\textwidth]{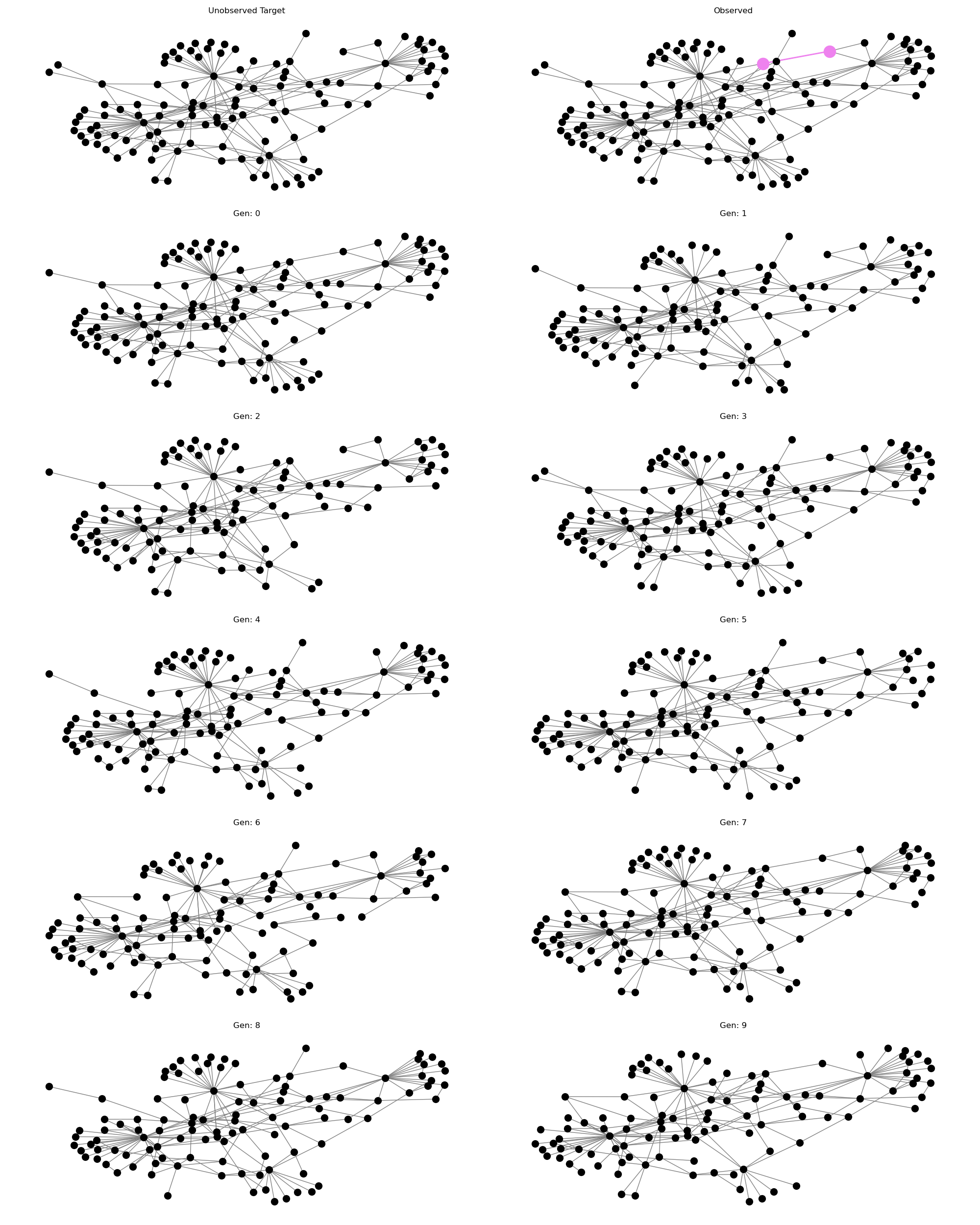}
    \includegraphics[width=0.3\textwidth]{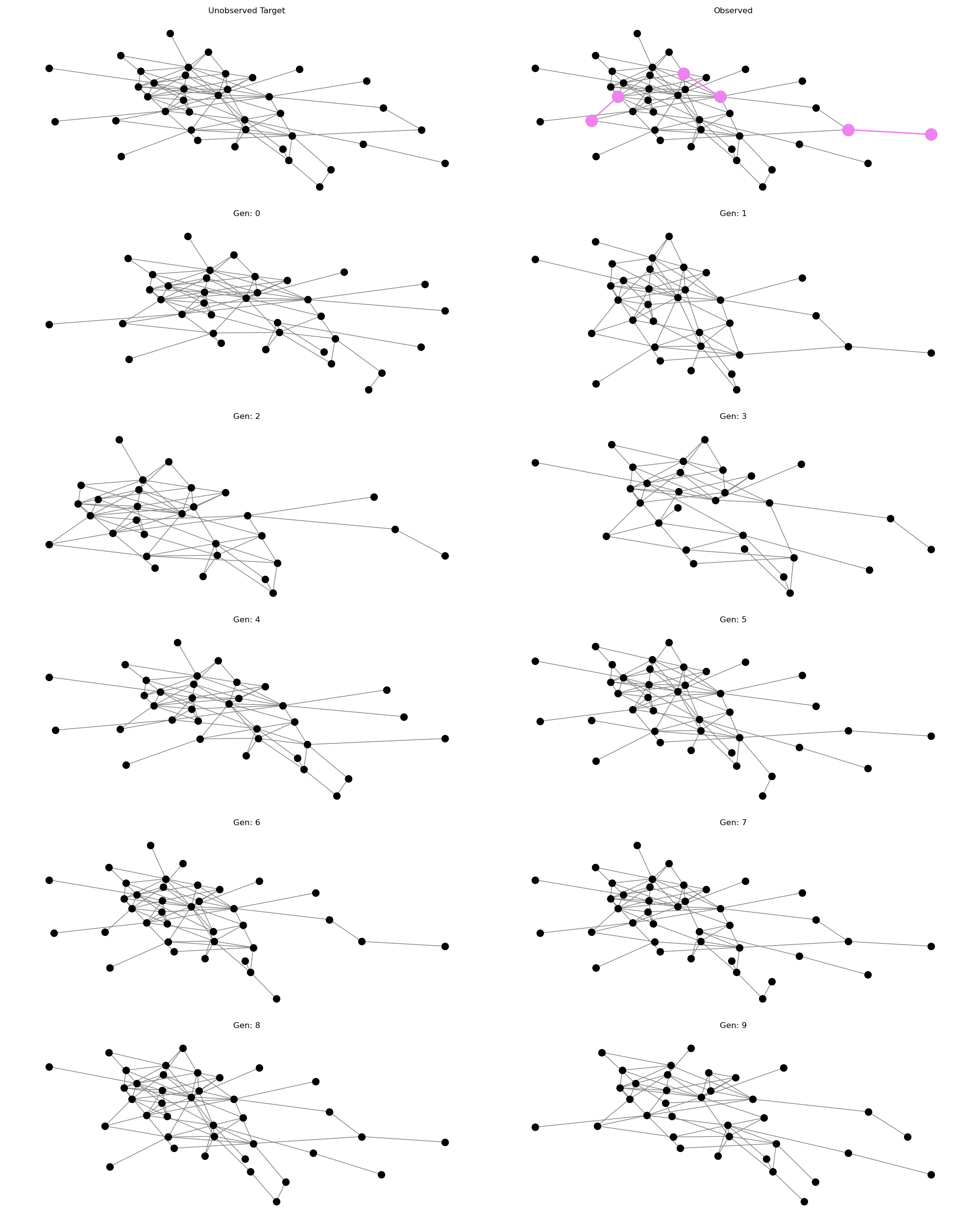}
    \label{fig:digress_bashapes_denoising_ex}
    \caption{\textbf{EDGE, (BAShapes, Cora, PolBlogs) Denoising.}}
\end{figure}





\end{document}